\documentclass{article}

\usepackage[preprint,nonatbib]{neurips_2024}

\usepackage{amsmath,amsfonts,bm}

\usepackage{tcolorbox}

\usepackage{xcolor,amsmath}
\usepackage[linesnumbered,ruled,vlined]{algorithm2e}
\DontPrintSemicolon

\usepackage{xcolor}
\definecolor{mygreen}{HTML}{3cb44b}
\usepackage{varwidth}

\SetKwComment{Comment}{\color{green!50!black}\# }{}

\SetKwProg{Function}{def}{:}{}

\SetKwProg{For}{for}{:}{}
\SetKwProg{If}{if}{:}{}
\newcommand{\VarSty}[1]{\textnormal{\ttfamily\color{blue!90!black}#1}\unskip}

\def\eqref#1{equation~\ref{#1}}

\def\1{\bm{1}}

\DeclareMathAlphabet{\mathsfit}{\encodingdefault}{\sfdefault}{m}{sl}
\SetMathAlphabet{\mathsfit}{bold}{\encodingdefault}{\sfdefault}{bx}{n}

\usepackage[utf8]{inputenc} 
\usepackage[T1]{fontenc}    
\usepackage{hyperref}       
\usepackage{url}            
\usepackage{booktabs}       
\usepackage{amsfonts}       
\usepackage{nicefrac}       
\usepackage{microtype}      
\usepackage{xcolor}         

\usepackage{minitoc}

\usepackage{multirow,tabularx}
\usepackage{makecell}
\usepackage{caption}
\usepackage{subcaption}
\usepackage{color, colortbl}
\usepackage{graphicx}
\usepackage{enumitem}
\usepackage{wrapfig}
\usepackage{tcolorbox}
\usepackage{CJKutf8}
\usepackage{algpseudocode}
\usepackage{amsmath} 

\definecolor{Gray}{gray}{0.93}

\definecolor{uclagold}{rgb}{1.0, 0.7, 0.0}
\definecolor{airforceblue}{rgb}{0.36, 0.54, 0.66}
\definecolor{rosegold}{rgb}{0.72, 0.43, 0.47}
\definecolor{pastelbrown}{rgb}{0.51, 0.41, 0.33}
\definecolor{isabelline}{rgb}{0.96, 0.94, 0.93}

\definecolor{macaroniandcheese}{rgb}{0.98, 0.89, 0.83}
\definecolor{wildblueyonder}{rgb}{0.85, 0.89, 0.95}

\definecolor{mediumtaupe}{rgb}{0.4, 0.3, 0.28}

\definecolor{bluegray}{rgb}{0.4, 0.6, 0.8}
\definecolor{celestialblue}{rgb}{0.29, 0.59, 0.82}
\definecolor{darkorange}{rgb}{1.0, 0.55, 0.0}

\definecolor{cadmiumred}{rgb}{0.89, 0.0, 0.13}
\definecolor{magnolia}{rgb}{0.97, 0.96, 1.0}
\definecolor{pastelblue}{rgb}{0.68, 0.78, 0.81}
\definecolor{persiangreen}{rgb}{0.0, 0.65, 0.58}
\definecolor{steelblue}{rgb}{0.27, 0.51, 0.71}

\definecolor{bluebell}{rgb}{0.64, 0.64, 0.82}
\definecolor{dimgray}{rgb}{0.41, 0.41, 0.41}
\definecolor{splashedwhite}{rgb}{1.0, 0.99, 1.0}

\definecolor{lavendergray}{rgb}{0.77, 0.76, 0.82}
\definecolor{lightgray}{rgb}{0.83, 0.83, 0.83}
\definecolor{lavendermist}{rgb}{0.9, 0.9, 0.98}

\definecolor{lightgreen}{HTML}{f8fcf4}
\definecolor{lightblue}{HTML}{dfebf7}

\newcommand{\llms}{LLMs}
\newcommand{\llm}{LLM}
\newcommand{\llmsfull}{Large Language Models}
\newcommand{\mllms}{MLLMs}
\newcommand{\mllmsfull}{Multimodal Large Language Models}

\newcommand{\lexvi}{\textsc{Seeker}}
\newcommand{\lexvit}{\textsc{Seeker-Tiny}}
\newcommand{\lexvis}{\textsc{Seeker-7B}}

\definecolor{splashedwhite}{rgb}{1.0, 0.99, 1.0}

\title{From Text to Pixel: Advancing Long-Context Understanding in MLLMs}

\author{Yujie Lu\textsuperscript{$^\spadesuit{}$}  
\quad Xiujun Li\textsuperscript{$^\heartsuit{}$} 
\quad Tsu-Jui Fu\textsuperscript{$^\spadesuit{}$}
\quad Miguel Eckstein\textsuperscript{$^\spadesuit$}
\quad William Yang Wang\textsuperscript{$^\spadesuit$}\\
\textsuperscript{$\spadesuit$}University of California, Santa Barbara \quad \textsuperscript{$\heartsuit$}University of Washington \quad \\
\url{https://github.com/YujieLu10/Seeker}
}

\begin{document}

\maketitle

\doparttoc 
\faketableofcontents 
\begin{center}
    \includegraphics[width=0.51\textwidth]{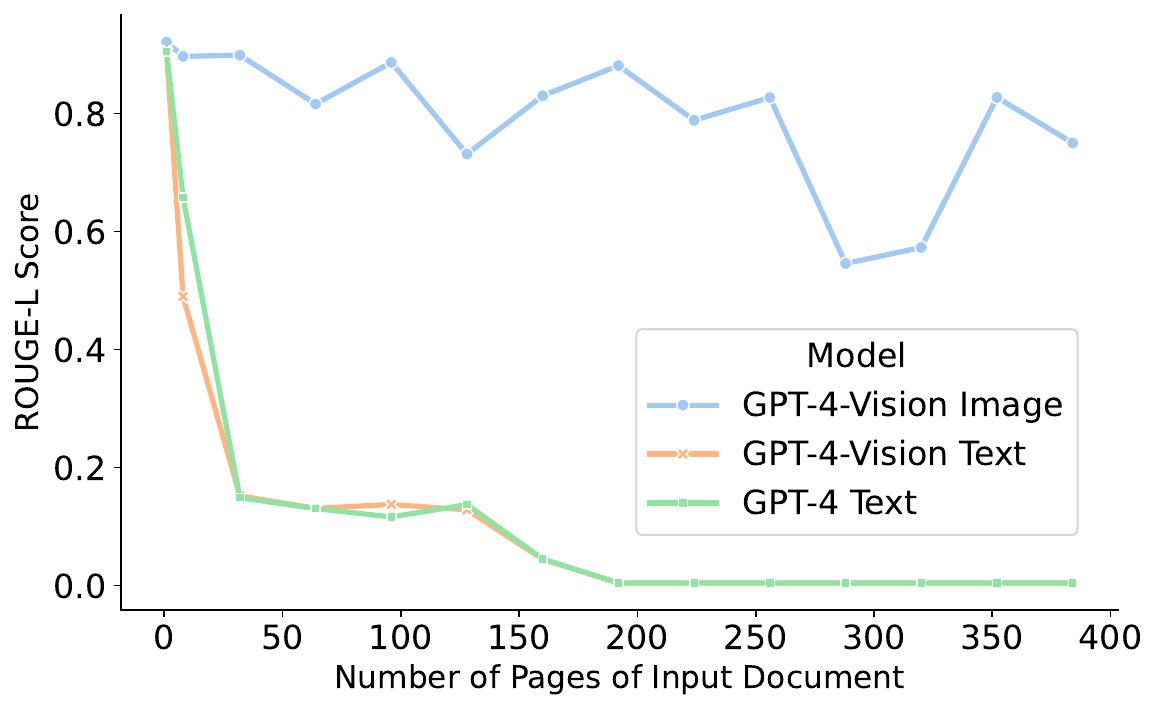}
    \includegraphics[width=0.48\textwidth]{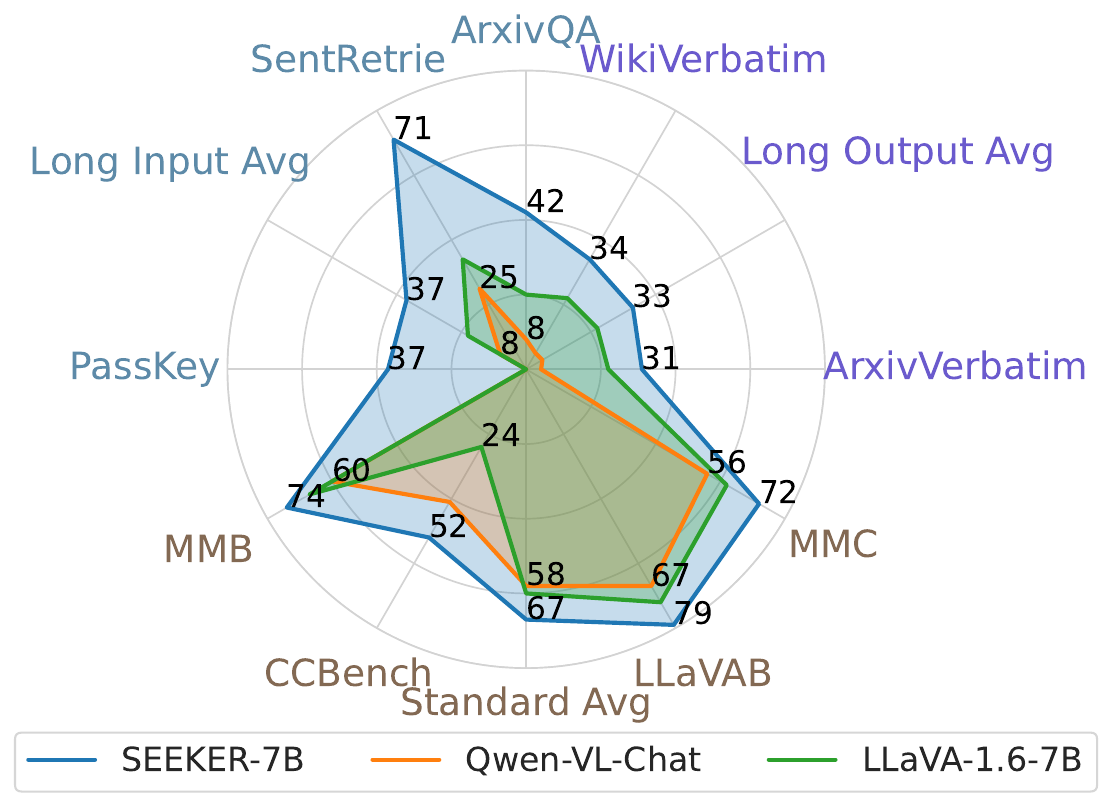}
    \captionof{figure}{Left: Performance plot on First-Sentence-Retrieval task revealing compact nature of image tokens in representing long content. Right: Radar chart demonstrating the superior performance of the SEEKER (ours) model across both short and long-context multimodal tasks.
    }
    \label{fig:teaser}
\end{center}%

\begin{abstract}
The rapid progress in \mllmsfull{} (\mllms) has significantly advanced their ability to process and understand complex visual and textual information. However, the integration of multiple images and extensive textual contexts remains a challenge due to the inherent limitation of the models' capacity to handle long input sequences efficiently. In this paper, we introduce \lexvi, a multimodal large language model designed to tackle this issue. \lexvi{} aims to optimize the compact encoding of long text by compressing the text sequence into the visual pixel space via images, enabling the model to handle long text within a fixed token-length budget efficiently. Our empirical experiments on six long-context multimodal tasks demonstrate that \lexvi{} can leverage fewer image tokens to convey the same amount of textual information compared with the OCR-based approach, and is more efficient in understanding long-form multimodal input and generating long-form textual output, outperforming all existing proprietary and open-source \mllms{} by large margins. 
\end{abstract}

\section{Introduction}
The success of \llmsfull{} (\llms)~\cite{chatgpt2022,touvron2023llama2,bai2023qwen,deepseekai2024deepseek} has significantly impacted various fields, notably \mllmsfull{} (\mllms{})~\cite{gpt4-v,liu2023llava,bai2023qwenvl,lu2024deepseekvl}. And there is a burgeoning interest in enhancing \llms{} to handle longer context~\cite{xiong2023effective,chen2024longlora,jin2024llm}, for example, the recent GPT-4O~\cite{gpt-4o} can support up to 128k tokens, paving the way to unlock many real-world applications from long-document understanding, summarization to document translation, among others.

\begin{figure*}[t!]
    \centering
    \includegraphics[width=\linewidth]{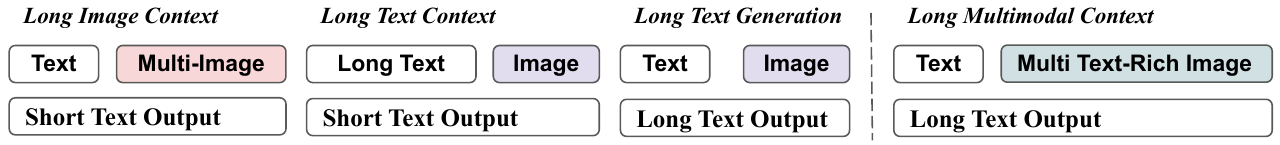}
    \caption{Long Multimodal Context Task mainly consists of two elements: 1) long image sequence and text input and 2) long text output.
    }
    \label{fig:longcontext_paradigm}
\end{figure*}

In many applications involving long-form documents that integrate images and text, there is a significant demand for the strong long-context understanding ability of \mllms. As shown in Figure~\ref{fig:longcontext_paradigm}, the long context in the multimodal domain falls into two main categories: 1) long-form inputs consisting of multiple text-rich images, and 2) long-form text outputs. In the first category, multiple images increase the context length with image tokens and additional text tokens if the images are text-rich. This requires the model to efficiently integrate textual data with multiple images and reason across them. In the second category, the model must produce coherent and attentive long responses to the input context, avoiding irrelevant or hallucinated content and minimizing reliance on the model knowledge without considering the specific multimodal context.

The existing \mllms{}~\cite{liu2023llava,liu2023improved,lu2024deepseekvl} leverage pretrained LLMs~\cite{vicuna2023,touvron2023llama} and inherit their advanced language understanding capabilities. Although these \mllms{} demonstrate strong performance across various vision-language benchmarks~\cite{liu2024mmbench,yu2023mmvet}, their effectiveness in long-form multimodal contexts is less explored. This issue becomes significant in tasks with very long input or output, which may exceed the context length limit (e.g., $2048$ tokens for LLaMA) and increase computational overhead.

While only a few \mllms{}~\cite{gpt4-v,mckinzie2024mm1} are capable of handling multiple images in the multimodal context, efficiency emerges as another critical challenge. ``A picture is worth a thousand words'', for human, it is more natural to fully utilize our bandwidth to process an image than words. However, this might not be the case for models. In this paper, we aim to represent information in a more compact form, enabling conveying more information within the same context length. Specifically, we investigate the ``visual token representation'' as an alternative to text tokens, and introduce \lexvi{}, an efficient method for managing long contexts within a constrained length budget. This approach allows us to process more context within a fixed token length.

\begin{figure*}[t!]
    \centering
    \includegraphics[width=\linewidth]{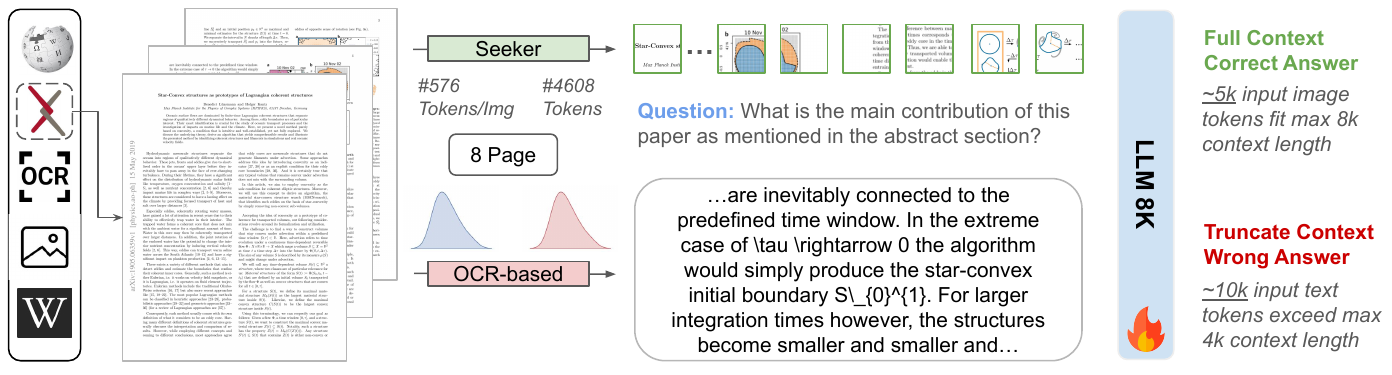}
    \caption{Our ~\lexvi~ surpass OCR-based model on long multimodal context tasks: 1) process multiple text-rich images naturally. 2) more compact token and fit easily in fix-context length LLM.
    }
    \label{fig:longcontext_mllm}
\end{figure*}

As shown in Figure~\ref{fig:longcontext_mllm}, an OCR-based approach might yield $10k$ tokens from an eight-page document for the \llm{} with a context limit of $8k$ tokens. While, \lexvi{} processes each of the eight pages as separate images, converting them into 576 tokens each. This generates a total of $4,608$ tokens for the whole document, which are then fed into the \lexvi{} model for reasoning and generation.

To the best of our knowledge, \lexvi{} is the first to address this in the long-context \mllms{} by employing a compact tokenization strategy that leverages visual tokens for textual information, thus reducing the number of tokens required and enabling the processing of longer texts without additional computation overhead. \lexvi's design allows for sophisticated reasoning across multiple images. By interleaving image tokens with textual data, \lexvi{} can preserve context coherence and continuity across extended sequences, enabling more effective interpretation and integration of visual data in scenarios where traditional text-based models may struggle. To sum up, our main contributions are as follows:
\begin{itemize}[itemsep=0pt,topsep=2pt,leftmargin=20pt] 
    \item We present \lexvi, an approach to leverage the visual tokens to process long documents more efficiently than OCR text tokens, given the same token length constraint.
    \item \lexvi{} supports long-context multimodal reasoning, effectively handling long-form multi-image input and generating long-form text output.
    \item Our instruction-tuned \lexvi{} model demonstrates promising results compared to the existing \mllms{} on six long-context multimodal tasks.
\end{itemize}

\section{Background}
\noindent \paragraph{Multimodal Large Language Model}
Recent advancements of proprietary \llmsfull, GPT-4~\cite{gpt4}, Gemini~\cite{team2023gemini}, Claude, QWen~\cite{bai2023qwen}, and open-source ones, LLaMA~\cite{touvron2023llama,touvron2023llama2}, Mistral, have shown groundbreaking applications. 
Their counterparts in the visual domain are followed up, including GPT-4V~\cite{gpt4-v}, Gemini-Vision~\cite{team2023gemini}, Claude3-Opus-VL, Qwen-VL~\cite{bai2023qwenvl}, InstructBLIP~\cite{dai2023instructblip}, LLaVA~\cite{liu2023visual}. Some work~\cite{lu2023vim,wu2024comprehensive} reveals the deficit of these \mllms~ in multiple images reasoning, and recent models~\cite{mckinzie2024mm1,laurençon2024matters,jiang2024mantis} improve such capabilities. Other work\cite{rust-etal-2023-pixel,gao2024improving} explore to process both text and images within pixels via task-specific finetuning. However, the long-context capabilities of these \mllms{} are underexplored. Our proposed \lexvi{} advances the long-context multimodal understanding of \mllms{} from two aspects, long-form image inputs and long-form text outputs.

\noindent \paragraph{Long Context Transformer}
The Transformer-dominated \llms{} have struggled with long context length as studied in \cite{liu2023lost}. LongLLaMA~\cite{tworkowski2023focused}, Self-Extend~\cite{jin2024llm} have been proposed to increase the effective context length by either fine-tuning or training-free approach based on pre-trained \llms~.
When it comes to \mllms{}, additional long-context issues are introduced from Vision Transformers (ViTs)~\cite{dosovitskiy2021image} for image processing, and connecting with the \llms{}. The concept of Dynamic Tokens~\cite{wang2021images} introduces a novel approach where the allocation of computational resources is adapted dynamically, emphasizing that not all image parts equally contribute to the recognition task. Additionally, the development of the Self-slimmed Vision Transformer~\cite{zong2022selfslimmed} introduces a mechanism for model slimming during the inference phase, reducing computational overhead without significant loss in accuracy.
In contrast, our proposed \lexvi{} utilizes image tokens as compact representations for image and text, alleviating the context length required for the same amount of semantic information in the language model backbone when processing multimodal content.

\section{\lexvi{}: Long-context Vision and Language Understanding}
We propose \lexvi{}, a multimodal large language model designed to handle long-context images and texts, as depicted in Figure~\ref{fig:longcontext_mllm}. In Section~\ref{sec:compact_text_repr}, we discuss the innovative use of image tokens to represent lengthy textual data compactly. Then we introduce long-context multimodal task and instruction data in Section~\ref{sec:long_context_multimodal_instruct}. Finally, in Section~\ref{sec:longcontext_mllm}, we illustrate the architecture of our \lexvi{} to support both long-context and short-context multimodal understanding.

\subsection{Using Image Tokens to Encode Text Helps Context Length Extrapolation}
\label{sec:compact_text_repr}
We follow the approach outlined in \cite{xiong2023effective} to evaluate model's extrapolation capability in the First-Sentence-Retrieval task. In this task, models are required to retrieve the first sentence at a specific length. We conduct this synthetic task on various numbers of documents with different page counts. We probe the performance of GPT-4-Vision Image by feeding its images of documents and compare it with GPT-4-Vision Text and GPT-4, which receive extracted text using the OCR model Nougat~\cite{blecher2023nougat}. Nougat achieves over a $90$ BLEU score on OCR text from scientific documents. All these models have a context length limit of $128k$ tokens.

On the left side of Figure~\ref{fig:teaser}, we visualize the Rouge-L~\cite{lin-2004-rouge} score in relation to the total number of pages of input documents, which range from $1$ (approximately $1k$ text tokens) to $448$ (approximately $500k$ text tokens). We observe a significant performance degradation in models fed with text input. In contrast, without any additional changes, we see improved extrapolation when representing length text content with visual tokens by feeding images of documents directly to the model.

\begin{table}[t]
\begin{center}
\caption{\textbf{Long-Context Multimodal Task.} \texttt{Img/\#In}: the number of input images, \texttt{Text Tok/\#In} and \texttt{\#Out}: the number of input and output text tokens. Full examples are presented in Appendix~\ref{sec:longcontext_task_examples}.
}
\resizebox{\columnwidth}{!}{
\begin{tabular}{l@{\hspace{5pt}}c|@{\hspace{5pt}}ccc}
\toprule
\multirow{2}{*}{Task}& \multirow{2}{*}{Prompt Example} & \multicolumn{1}{c}{\texttt{Img}} & \multicolumn{2}{c}{\texttt{Text Tok}} \\
\cmidrule{3-3}\cmidrule{4-5}
   & & \#In. & \#In. & \#Out. \\ 
\midrule
\multicolumn{5}{c}{\textit{Long-Form Multi-Image Input}} \\
\midrule
\texttt{Index} & Which Image contains the given sentence? & $6.6$	& $100.4$ &	$1.0$ \\
\texttt{SentRetrie} & What is the first sentence on the first image? & $1.0$	& $23.0$ &	$35.5$ \\
\texttt{ArxivQA} & What is the main purpose of the article as stated in the abstract? & $8.2$	& $13.9$ &	$35.0$ \\
\texttt{PassKey} & What is the <PASSKEY> in the provided images? & $4.0$	& $95.4$ &	$2.6$ \\
\midrule
\multicolumn{5}{c}{\textit{Long-Form Text Output}} \\
\midrule
\texttt{ArxivVerb} & Read the text in the image verbatim. & $1.0$	& $10.0$ &	$1301.6$ \\
\texttt{WikiVerb} & Read the text in the image verbatim. & $1.0$	& $16.0$ &	$1107.1$ \\
\bottomrule
\end{tabular}
}

\label{tab:longcontext_task}
\end{center}
\vspace{-4mm}
\end{table} 

\subsection{Long-Context Multimodal Task}
\label{sec:long_context_multimodal_instruct}
We mainly consider two categories of long-context multimodal capabilities, as outlined in Table~\ref{tab:longcontext_task}: 1) Long-form multimodal input: This involves multiple text-rich images interleaved with text as the input context. 2) Long-form text output: This requires generating long text.

\noindent \paragraph{Instruction Data for Long-Form Multi-Image Input}
First, we combine an arbitrary number of single-image visual instruction data~\cite{liu2023llava} sourced from CC3M into the multi-image format for the intra-image reasoning task. This helps initiate model's capability of understanding sequences of images (e.g., \textit{<$img_{1}$> This image depicts a... <$img_{2}$> This image shows a...}). We then curate inter-image reasoning instruction data from NLVR2~\cite{suhr-etal-2019-corpus} (e.g., \textit{<$img_{1}$> <$img_{2}$> Considering the images on both sides, is `At least one of the televisions is turned off.' valid? Answer yes or no.}), Mimic CGD (e.g., \textit{<$img_{1}$> <$img_{2}$> What's the difference between the two sinks in the images?}), and annotate multi-image conversation data on COCO images~\cite{lin2015microsoft} using GPT-4V (e.g., \textit{<$img_{1}$> <$img_{2}$> <$img_{3}$> How many birds are in all the provided images?}).
To enable understanding of long-form text-rich image sequences, we collect compiled PDFs from arXiv documents. Each page from these documents is processed as images, ranging from 4 to 24 pages. We use GPT-4V to generate descriptive or conversational instruction data for these scientific documents.
To further improve the model's understanding of each provided image, we create a multi-image text grounding task, requiring the model to ground the question to the referred image (e.g., \textit{<$img_{1}$> <$img_{2}$> ... <$img_{8}$> Which image contains the answer to the question / Which image contains the sentence...}).
\noindent \paragraph{Instruction Data for Long-Form Text Output}
To enhance long-form text generation capabilities related to the given image, we propose a task that involves reading the text in the image verbatim (e.g., \textit{<$img_{1}$> Quote the text in the image verbatim.}). This challenging task requires the vision backbone to encode character-level image details and the language backbone to attend to the image token while producing very long text without hallucinating on previously generated content.

\subsection{Long-Context Multimodal Large Language Model}
\label{sec:longcontext_mllm}
To enable long-context multimodal reasoning, our model architecture should: 1) encode multiple images interleaved with text, 2) align images and text at a fine-grained level, and 3) decode long texts that attend to extended multimodal contexts. The following paragraphs illustrate the design of our proposed \lexvi~ for this purpose.

\noindent \paragraph{Long-Context Multi-Image Encoding}
For effective feature integration in scenarios involving multiple images, it is crucial to include image separators to concatenate text and image sequences as:
\begin{equation}
\begin{aligned}
\text{Query} = \text{Query}_{\text{system}} + \sum_{i=1}^N \left( \mathbf{Q}_{\text{img},i} + \mathbf{Q}_{\text{txt},i} \right)\\
\mathbf{Q}_{\text{img},i} = \text{start}(\text{img},i) + \text{content}(\text{img},i) + \text{end}(\text{img},i)
\end{aligned}
\end{equation}
Specifically, we use \text{start}(\text{img},i) and \text{end}(\text{img},i) as special tokens `<|startofimgi|>' and `<|endofimgi|>' to distinguish the start and end of each image, respectively.
We observe this strategy is essential for maintaining model performance, especially when training is limited to a small dataset of long-context multimodal instructions.
The encoding process and the concatenation of the feature vectors of the input sequence can be described as:
\begin{equation}
\begin{aligned}
t_i = \mathrm{Enc_{t}(T_i)}, v_i = \mathrm{MLP_{v\rightarrow t}(Enc_{v}(I_i))} \\
\quad Q = [t_0; v_1; t1; v_2; t2; \ldots; v_n; tn]
\end{aligned}
\end{equation}
Here, $Enc_{v}$ encodes each image $i$ into a feature vector and projects it to the word embedding space. The concatenated vector $Q$ integrates sequences of image and text feature vectors, where $[;]$ denotes concatenation along the feature dimension.

Additionally, to preserve the model's capability with single-image data without necessitating re-finetuning, we introduce image-specific identifiers only during multi-image training and inference, while retaining the original prompt template for single-image contexts. Furthermore, incorporating image-index-aware question-answering instruction data enhances the model's ability to anchor its reasoning to specific images, enabling robust multi-image understanding and reasoning.

\noindent \paragraph{Dense Image-Text Alignment}
We inherit the general image-text alignment from the pre-training image-text pairs.
To enhance the visual representation of dense text in images, and improve the alignment between image and text representation of rendered text, we curate a visual-embedded task that renders text into visual space.
Specifically, we render text paragraphs from Wikipedia into $1024 \times 1024$ images using Arial font, with sizes ranging from 18 to 30, providing various word densities per image. We observe that it is essential to start by learning image-text alignment at a sparse level (large font size, low word density) and gradually incorporate dense text-rendered image data.
Task types we consider include question answering on multiple images rendered with text from Wikipedia, and reading the text verbatim from rendered images.

\noindent \paragraph{Supervised Fine-tuning Strategy}
We aim to leverage sequential data processing to fine-tune models on a combination of textual and visual inputs, enabling them to generate coherent and contextually relevant responses based on both text and image data. In the domain of multimodal large language models, the autoregressive training objective is a pivotal technique, which can be formulated as follows:
\begin{equation}
\begin{aligned}
p(X_{o}|Q) = \prod_{i=1}^{L} p_{\theta}(x_{i}|Q)\\
\mathcal{L}(\theta) = -\sum_{t=1}^{L} \log P(x_i | x_{<i}, Q; \theta)
\end{aligned}
\end{equation}
where $x_i$ represents tokens with length $L$, $X_O$ denotes the target output given the features of multimodal queries $Q$, and $\theta$ denotes the model parameters. This loss function encourages the model to predict the next token in the sequence, given the previous visual and textual tokens.

\section{Implementation Details}
\subsection{Model Architecture}
The language model backbone of \lexvi{} is the DeepSeek LLM~\cite{deepseekai2024deepseek}, which has a design similar to LLaMA. It is supervised-finetuned on 2T tokens with additional DPO and surpasses LLaMA-2 and GPT-3.5 on numerous open-eval tasks.
To enable to process high-resolution images and ensure adept performance in real-world scenarios, we instruction-tune the stage-3 model from the DeepSeek-VL series of model~\cite{deepseekai2024deepseek}.
The vision encoder of \lexvit{} is SigLIP, and the vision encoder of \lexvi{} is a hybrid of SigLIP-L~\cite{zhai2023sigmoid} and SAM-B~\cite{kirillov2023segment}. This enables processing $1024 \times 1024$ images into a fixed token length of $576$. This fixed token length for high-resolution image processing provides an optimal balance of fine-grained and compact visual representation. The adaptor used is a hybrid MLP, the same as in DeepSeek-VL~\cite{lu2024deepseekvl}.

\subsection{Training}
We use the AdamW~\cite{loshchilov2019decoupled} optimizer to train our models for 1 epoch with a batch size of 32. The learning rate is linearly warmed up during the first $5\%$ of steps to $1e-4$ and then reduced to zero using a cosine learning rate scheduler. The context sequence length is set to 4096 during instruction-tuning on single-image data. For continual training on our proposed long-context multimodal instruction data (Section~\ref{sec:long_context_multimodal_instruct}), we set the maximum length to 8192 to accommodate a long sequence of images and long-form text output.
We set the rank to 8 for low-rank adaptation (LoRA~\cite{hu2021lora}). Our \lexvi~ and \lexvit~ are trained on a single 8-A100-40G node for 30 hours and 12 hours, respectively.

\subsection{Evaluation}
Details of each long-context multimodal task are introduced in Table~\ref{tab:longcontext_task}, with more details presented in Appendix~\ref{sec:longcontext_task_examples} and Appendix~\ref{sec:evaluation_details}. Each long-context multimodal task contains $80$ diversified samples. We use the accuracy metric for the multiple-choice task (\texttt{Index}) and the Rouge-L score for all other text generation tasks.
For standard multimodal tasks, which require fewer than four image inputs and text answers that are less than $400$ tokens.
We use the accuracy metric for multiple-choice NLVR2~\cite{suhr-etal-2019-corpus} test-public split and the BLINK~\cite{fu2024blink} validation split. We validate models on the official evaluation metrics and test splits for general single-image multimodal benchmarks (MMB EN, MMB CN(MMC) and Circular Eval for MMB (CCBench) ~\cite{liu2024mmbench}, SEED~\cite{li2023seedbench}, AI2D~\cite{Kembhavi2016ADI}, LLaVAB~\cite{liu2023llava}, ChartQA~\cite{masry2022chartqa}, TextVQA~\cite{singh2019vqa}). We follow the inference configurations in VLMEvalKit~\cite{2023opencompass}.

\begin{table}[t]
\begin{center}
\caption{\textbf{Long Image and Text Context}. \colorbox{isabelline}{\makebox(5,5){}}: proprietary models, \colorbox{lightblue}{\makebox(5,5){}}: the proposed models, \texttt{\#Tok/Img}: the number of tokens per image. We report accuracy on multiple-choice task \texttt{Index}, and Rouge-L score for other tasks.
}
\resizebox{\linewidth}{!}{
\begin{tabular}{lcc|cccccccc}
\toprule
\multirow{2}{*}{Models} & \multirow{2}{*}{Params} & \multirow{2}{*}{\#Tok/Img}  & \multicolumn{5}{c}{\texttt{Long-Form Multi-Image Input}}  & \multicolumn{3}{c}{\texttt{Long-Form Text Output}}  \\
\cmidrule(lr){4-8}\cmidrule(lr){9-11}
 &  &  & \texttt{Index} &  \texttt{SentRetrie} & \texttt{ArxivQA} & \texttt{PassKey} & \texttt{Avg} & \texttt{ArxivVerb} & \texttt{WikiVerb} & \texttt{Avg} \\
 \midrule
\multicolumn{1}{l}{\textbf{Close-source MLLMs}} \\
\rowcolor{isabelline} GPT-4V~\cite{gpt4-v} & $-$ & $85$ & $32.50$ & $71.10$ & $45.19$ & $27.16$ & $43.98$ & $32.58$ & $5.96$ & $19.27$ \\
\midrule
\multicolumn{1}{l}{\textbf{Open-source MLLMs}} \\
Qwen-VL-Chat~\cite{bai2023qwenvl} & $7B$ & $-$ & $2.49$ & $25.05$ & $8.24$ & $0.00$ & $8.94$ & $4.90$ & $5.41$ & $5.15$ \\
LLaVA-1.5~\cite{liu2023improvedllava} & $7B$ & $576$ & $23.74$ & $30.61$ & $35.60$ & $0.00$ & $22.48$ & $4.14$ & $3.80$ & $3.97$ \\
LLaVA-Next~\cite{liu2024llavanext} & $7B$ & $2880$ & $17.49$ & $34.35$  & $20.50$ & $0.00$& $18.08$ & $22.33$ & $22.94$ & $22.63$ \\
LLaVA-Next (Mistral)~\cite{liu2024llavanext} & $7B$ & $2880$ & $17.49$ & $34.45$ & $21.39$ & $0.00$ & $18.33$ & $20.11$ & $20.92$ & $20.51$ \\
DeepSeek-VL~\cite{lu2024deepseekvl} & $7B$ & $576$ & $13.74$ & $10.37$ & $19.83$ & $0.17$ & $11.02$ & $\underline{31.59}$ & $16.48$ & $24.03$ \\
IDEFICS2~\cite{laurençon2024matters} & $8B$ & $64$ & $10.83$ & $63.46$ & $9.68$ & $0.13$ & $21.02$ & $12.12$ & $5.93$ & $9.02$ \\
Monkey-Chat~\cite{li2023monkey} & $10B$ & $-$ & $16.24$ & $23.65$ & $17.90$ & $0.00$ & $14.44$ & $5.82$ & $2.08$ & $3.95$ \\
LLaVA-1.5~\cite{liu2023improved} & $13B$ & $576$ & $22.49$ & $41.02$ & $32.31$ & $0.00$ & $23.95$ & $9.57$ & $7.12$ & $8.34$ \\
LLaVA-Next~\cite{liu2024llavanext} & $13B$ & $2880$ & $11.24$ & $37.55$ & $15.60$ & $0.00$ & $16.09$ & $27.14$ & $\underline{31.05}$ & $\underline{29.09}$ \\
\midrule
\multicolumn{1}{l}{\textbf{Open-source Tiny MLLMs}} \\
DeepSeek-VL~\cite{lu2024deepseekvl} & $1.3B$ & $576$ & $14.99$ & $10.46$ & $21.29$ & $0.15$ & $11.72$ & $20.06$ & $10.43$ & $15.24$ \\
MiniCPM-V~\cite{hu2024minicpm} & $3B$ & $-$ & $8.74$ & $12.01$ & $31.42$ & $0.00$ & $13.04$ & $1.50$ & $2.98$ & $2.24$ \\
\midrule
\multicolumn{1}{l}{\textbf{Ours}} \\
\rowcolor{lightblue} 
\lexvit~ & $1.3B$ & $576$ & $\textbf{33.74}$ & $\underline{66.99}$ & $\textbf{42.68}$ & $\underline{24.99}$ & $\underline{42.10}$ & $23.52$ & $25.33$ & $24.42$ \\
\rowcolor{lightblue} 
\lexvi~ & $7B$ & $576$ & $\underline{27.49}$ & $\textbf{71.33}$ & $\underline{42.35}$ & $\textbf{37.91}$ & $\textbf{44.77}$ & $\textbf{31.85}$ & $\textbf{34.98}$ & $\textbf{33.41}$ \\

\bottomrule
\end{tabular}
}
\label{tab:mllm_long_bench}
\end{center}
\end{table}

\section{Main Results}
\subsection{Long Image and Text Context}
\noindent \paragraph{Long-Form Multi-Image Input}
In Table~\ref{tab:mllm_long_bench}, \lexvi~ achieves significantly surpassing larger open-source \mllms~ across all four long-form multi-image input tasks. We concatenate the images for models that can not handle image sequences.
Additionally, \lexvit~ ranks second best. On average, our models also outperform the proprietary GPT-4V model.
This indicates our auxiliary tasks, as detailed in Section~\ref{sec:long_context_multimodal_instruct}, enhance the models' reasoning across multiple images and grounding content to specific images. 
Thus our models excel at handling long-context tasks involving long-form multiple text-rich image inputs.

\noindent \paragraph{Long-Form Text Output}
In Table~\ref{tab:mllm_long_bench}, our \lexvi~ achieves the best performance for long-context tasks requiring long-form text output. On average, LLaVA-Next~\cite{liu2024llavanext}-13B also performs well, likely because these tasks usually require a single image. Its feature of splitting images into four tiles as additional 2304 image tokens, combined with the original image, greatly enhances its ability to capture visual details. This is particularly beneficial for verbatim tasks involving Arxiv and Wikipedia content rendered in the image.
Meanwhile, DeepSeek-VL~\cite{lu2024deepseekvl} achieves the best scores among other open-source 7B \mllms~, primarily due to its alignment of image and text by enforcing text reading from a large scale of visual-situated real-world data, such as documents and PDFs.
By incorporating our small-scale verbatim task data, which includes images rendered with text of various font sizes, into the instruction-tuning stage, our models achieve a $38.1\%$ performance improvement.

\noindent \paragraph{Fix-length Image Tokens are more Expressive than Text Tokens}
\begin{wraptable}{r}{8.5cm}
\begin{center}
\caption{Probing Question Answering with Varying Page Context: Our \lexvi~ model seeks more accurate text answers within compact image tokens of image sequences compared to OCR-based approaches with the same context length.}
\resizebox{.6\columnwidth}{!}{
\begin{tabular}{l@{\hspace{5pt}}c|@{\hspace{5pt}}ccccc}
\toprule
\multirow{2}{*}{Models}& \multirow{2}{*}{Input Type} & \multicolumn{5}{c}{\texttt{ArxivQA}} \\
   & & p=4:6  & p=6:8  & p=8:10  & p=10:12  & Avg \\ 
\midrule
\multicolumn{7}{c}{\textit{LLM}} \\
\midrule
DeepSeek-LLM & OCR Txt & $35.79$ & $35.74$ & $36.00$ & $29.99$ & $34.38$\\
\lexvi~-LLM & OCR Txt & $\textbf{45.26}$ & $46.17$ & $50.57$ & $39.18$ & $45.29$\\
\midrule
\multicolumn{7}{c}{\textit{MLLM}} \\
\midrule
DeepSeek-VL & Seq Img & $29.30$ & $37.97$ & $36.67$ & $28.38$ & $33.08$\\
\lexvi~ & Seq Img+OCR Txt & $35.30$ & $41.22$ & $40.73$ & $33.49$ & $37.68$ \\
\lexvi~ & Seq Img & $44.43$ & $\textbf{50.81}$ & $\textbf{58.10}$ & $\textbf{39.95}$ & $\textbf{48.32}$\\
\bottomrule
\end{tabular}
}
\label{tab:mllm_arxivqa_ablation}
\end{center}
\vspace{-4mm}
\end{wraptable}

If a model can interpret text within images, it confirms that this method is a valid way to present information. Additionally, if the model requires fewer image tokens than text tokens to understand the text, this indicates that pixels can represent text more compactly.
To investigate this, we conduct a probing task involving question-answering using various pages of documents fed into the model, as shown in Table~\ref{tab:mllm_arxivqa_ablation}.
Notably, in this task, we use a version of our \lexvi~ with the same context length as the compared model, which is 4,096 tokens.
Our observations indicate that when the text token count is up to around 4,000, the response accuracy remains within the context length limit of 4,096 tokens without performance degradation for the language model (LLM). When the text token count exceeds 4,000 but the image token count remains below 4,000, the vision-language model (VLM) outperforms the LLM by 4 to 8 percentage points. However, when the image token count exceeds 4,000, the performance of the VLM also declines, though it remains slightly superior to that of the LLM.

\begin{table}[t]
\begin{center}
\caption{\textbf{Short Image and Text Context}. \colorbox{isabelline}{\makebox(5,5){}}: proprietary models, \colorbox{lightblue}{\makebox(5,5){}}: the proposed models.
}
\resizebox{\linewidth}{!}{
\begin{tabular}{l|cccccccccccc}
\toprule
\multirow{2}{*}{Models} &  \multicolumn{3}{c}{\texttt{Multi-Image}}  & \multicolumn{8}{c}{\texttt{Single-Image}}  \\
\cmidrule(lr){2-4}\cmidrule(lr){5-13}
& NLVR2 & BLINK & Avg & MMB & MMC & SEED & CCBench & AI2D  & LLaVAB & ChartQA & TextVQA & Avg \\ \midrule
\multicolumn{1}{l}{\textbf{Close-source MLLMs}} \\
\rowcolor{isabelline} GPT-4V~\cite{gpt4-v} & $71.7$ & $51.1$ & $61.4$ & $75.1$ & $74.4$ & $71.6$ & $46.5$ & $75.9$ & $93.1$ & $78.5$ & $78.0$ & $60.3$\\
\midrule
\multicolumn{1}{l}{\textbf{Open-source MLLMs}} \\
Qwen-VL-Chat~\cite{bai2023qwenvl} & $30.8$ & $28.1$ & $29.5$ & $60.6$ & $56.3$ & $64.8$ & $41.2$ & $63.0$ & $67.7$ & $49.8$ & $60.7$ & $58.0$ \\
LLaVA-1.5-7B~\cite{liu2023improved} & $61.7$ & $37.1$ & $49.4$ & $65.2$ & $59.0$ & $65.8$ & $27.5$ & $55.5$ & $61.8$ & $17.8$ & $45.4$ & $49.8$ \\
LLaVA-Next-7B~\cite{liu2024llavanext} & $58.7$ & $41.2$ & $49.9$ & $67.4$ & $62.3$ & $69.6$ & $24.3$ & $67.0$ & $72.7$ & $55.4$ & $64.4$ & $60.4$ \\
LLaVA-Next-7B (Mistral)~\cite{liu2024llavanext} & $43.5$ & $37.5$ & $40.5$ & $69.5$ & $61.3$ & $\textbf{72.4}$ & $30.0$ & $\underline{69.0}$ & $67.8$ & $51.8$ & $65.2$ & $63.1$ \\
DeepSeek-VL-7B~\cite{lu2024deepseekvl} & $46.6$ & $40.9$ & $43.7$ & $\textbf{74.1}$ & $71.4$ & $70.4$ & $\underline{51.7}$ & $65.3$ & $77.8$ & $59.1$ & $64.9$ & $\underline{66.8}$ \\
IDEFICS2-8B~\cite{laurençon2024matters} & $\textbf{79.9}$ & $\textbf{46.8}$ & $\textbf{63.4}$ & $75.3$ & $67.3$ & $71.9$ & $37.6$ & $72.3$ & $49.1$ & $24.36$ & $68.9$ & $66.3$ \\
Monkey-Chat-10B~\cite{li2023monkey} & $66.0$ & $40.5$ & $53.3$ & $71.0$ & $65.8$ & $68.9$ & $48.4$ & $68.5$ & $60.5$ & $\underline{59.5}$ & $\underline{65.5}$ & $63.5$ \\
LLaVA-1.5-13B~\cite{liu2023improved} & $66.2$ & $\underline{42.7}$ & $54.4$ & $69.2$ & $65.0$ & $68.2$ & $30.4$ & $61.1$ & $66.1$ & $18.2$ & $48.9$ & $53.4$ \\
LLaVA-Next-13B~\cite{liu2024llavanext} & $64.3$ & $42.6$ & $53.4$ & $70.7$ & $\textbf{79.0}$ & $\underline{71.9}$ & $28.8$ & $\textbf{72.2}$ & $73.9$ & $\textbf{61.4}$ & $\textbf{66.9}$ & $65.6$ \\
\midrule
\multicolumn{1}{l}{\textbf{Open-source Tiny MLLMs}} \\
DeepSeek-VL-1.3B~\cite{lu2024deepseekvl} & $61.3$ & $38.8$ & $50.1$ & $64.0$ & $62.9$ & $66.0$ & $37.6$ & $51.5$ & $51.1$ & $47.4$ & $57.8$ & $54.8$ \\
MiniCPM-V-3B~\cite{hu2024minicpm} & $63.1$ & $40.0$ & $51.5$ & $67.9$ & $62.6$ & $65.6$ & $41.4$ & $56.3$ & $51.3$ & $44.2$ & $56.6$ & $55.7$ \\
\midrule
\multicolumn{1}{l}{\textbf{Ours}} \\
\rowcolor{lightblue} 
\lexvit~-1.3B & $69.9$ & $40.5$ & $55.2$ & $64.8$ & $63.7$ & $66.0$ & $37.3$ & $49.0$ & $\textbf{81.7}$ & $45.4$ & $56.3$ & $58.0$ \\
\rowcolor{lightblue}
\lexvi~-7B & $\underline{72.4}$ & $42.1$ & $\underline{57.2}$ & $\underline{74.0}$ & $\underline{72.6}$ & $71.1$ & $\textbf{52.0}$ & $64.6$ & $\underline{79.3}$ & $58.3$ & $65.3$ & $\textbf{67.1}$ \\
\bottomrule
\end{tabular}
}
\label{tab:mllm_short_bench}
\end{center}
\end{table}

\subsection{General Multimodal Understanding Benchmark}
We seek to test general multimodal understanding and reasoning capabilities of our model, compared with the state-of-the-art models.
In Table~\ref{tab:mllm_short_bench}, we compare the performance of various models on both multi-image and single-image general multimodal benchmarks.
Our \lexvi~ achieves on par performance on short-context multi-image tasks among models of similar size.
Furthermore, despite not including general single-image instruction data in our continual instruction tuning on long-context tasks, our model still maintains performance on par with other \mllms~, and even outperforms all other models in some tasks. This performance preservation, without the need for additional instruction tuning data, is primarily due to our use of a separate image identifier for multi-image processing while retaining the single-image template during inference.

\section{Analysis}
\subsection{Context Length Extrapolation}
\begin{wrapfigure}{r}{0.5\textwidth}
  \begin{center}
    \includegraphics[width=0.5\textwidth]{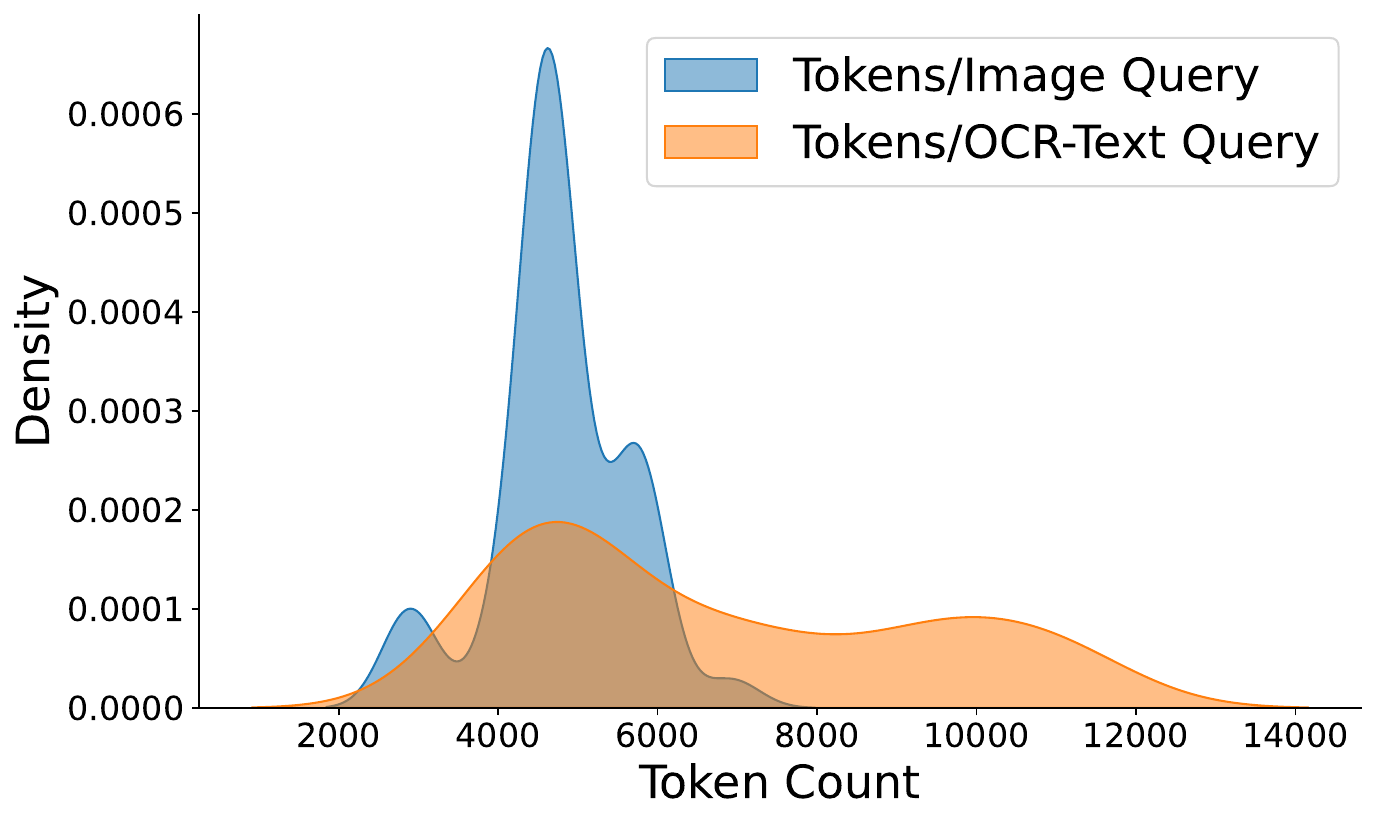}
  \end{center}
    \caption{
    Density plot comparing token counts for image token (blue) and OCR-text (orange) representations. Image tokens are more compact than text, fitting well within 8192 context length.}
  \label{fig:token_density_plot}
\end{wrapfigure}

We analyze the effectiveness of using image tokens versus OCR text tokens for image representation. The density plot in Figure~\ref{fig:token_density_plot} illustrates the distribution of token counts for both methods. The Image token representation is notably more compact, with a significant peak at lower token counts, whereas the OCR-text displays a broader distribution with higher counts. This variation shows that OCR-text length can be vulnerable and uncontrollable in images rich in text, often leading to wide-ranging token counts. In contrast, image tokens maintain a consistent token length regardless of textual density. With a model context length set to 8192 tokens, image tokens are handled 100\% of the time without truncation, whereas OCR-text frequently exceeds this limit, achieving only 66.25\% execution success without truncation. Meanwhile, truncating OCR text compromises performance as shown in Table~\ref{tab:mllm_arxivqa_ablation}. This highlights the advantages of image tokens for predictable and efficient encoding of long multimodal contexts.

\subsection{Inference Efficiency}
\begin{wrapfigure}{r}{0.5\textwidth}
  \begin{center}
    \includegraphics[width=0.5\textwidth]{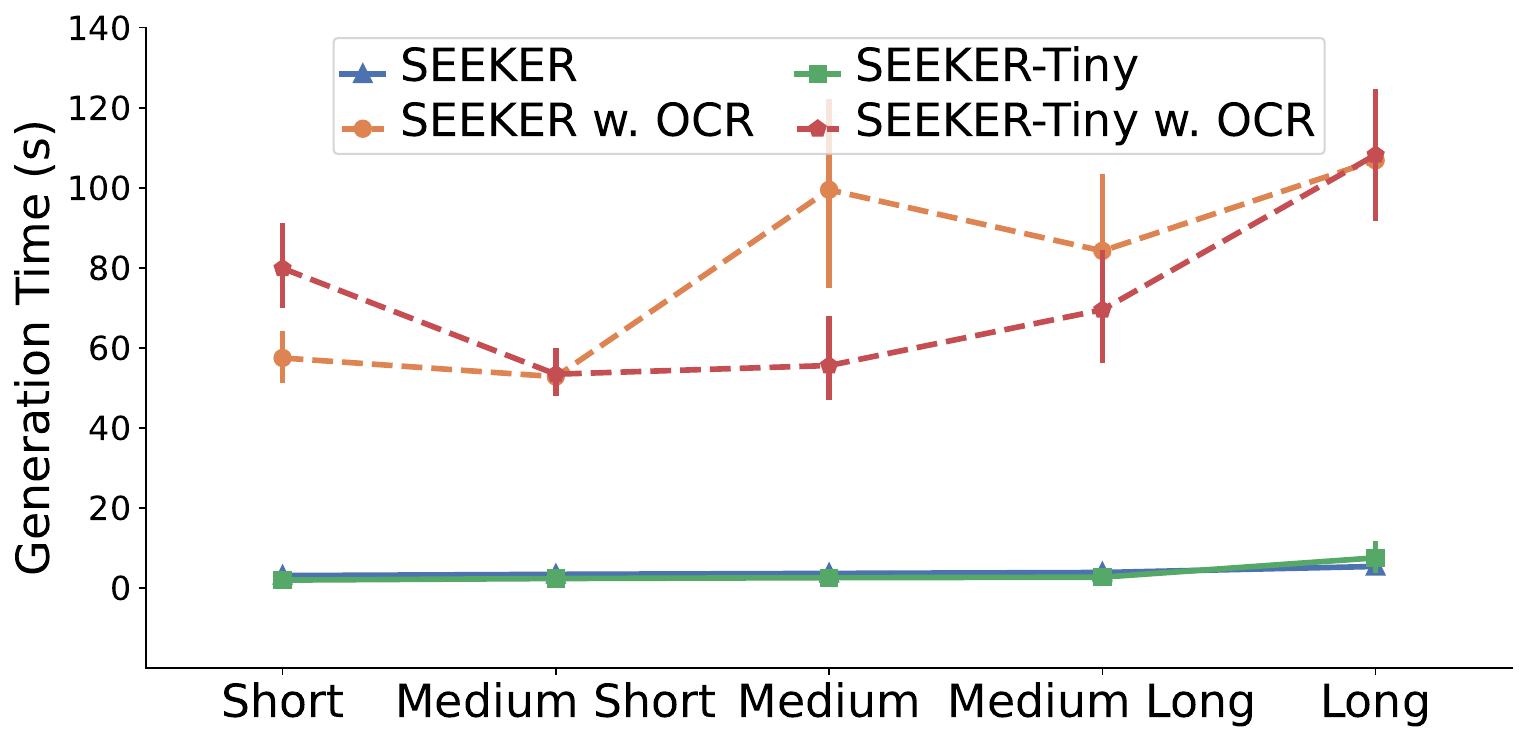}
  \end{center}
    \caption{
    Generation times for \lexvi~ and \lexvit~ with and without OCR.}
  \label{fig:generation_time}
\end{wrapfigure}
In addition to its context length extrapolation capability, our model \lexvi~ solves long-context multimodal tasks more efficiently compared to the OCR-based approach. For example, when comparing the inference time cost of \lexvi~ with and without OCR, the latter first extracts long text from multiple images and then feeds text into \lexvi~. By eliminating the time-consuming OCR step, our model achieves a significant reduction in inference time. Specifically, in the longest context scenario, \lexvi~ is approximately three times faster than OCR-based approach, showcasing the substantial time efficiency.

\subsection{Qualitative Showcases}
\begin{figure*}[t!]
    \centering
    \includegraphics[width=\linewidth]{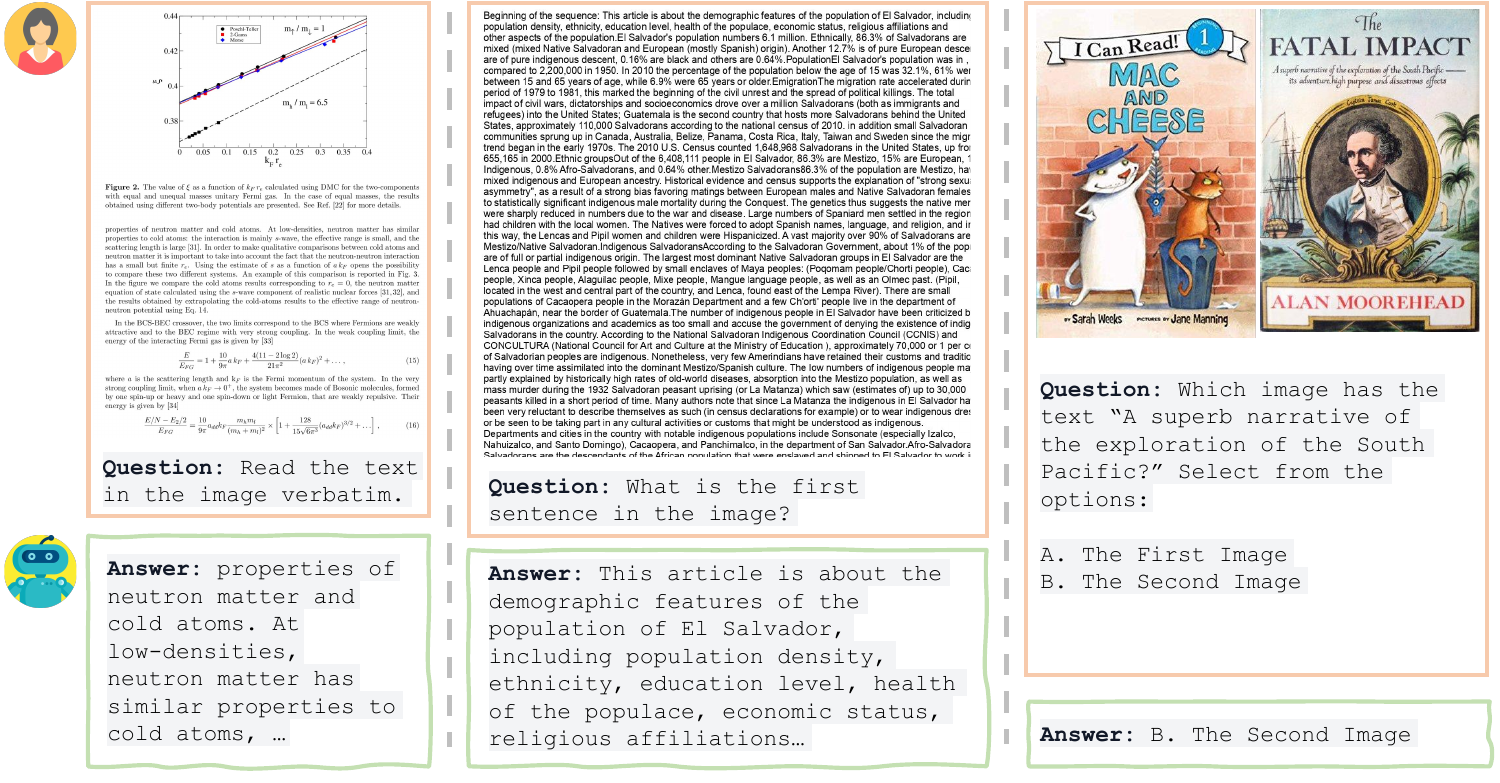}
    \includegraphics[width=\linewidth]{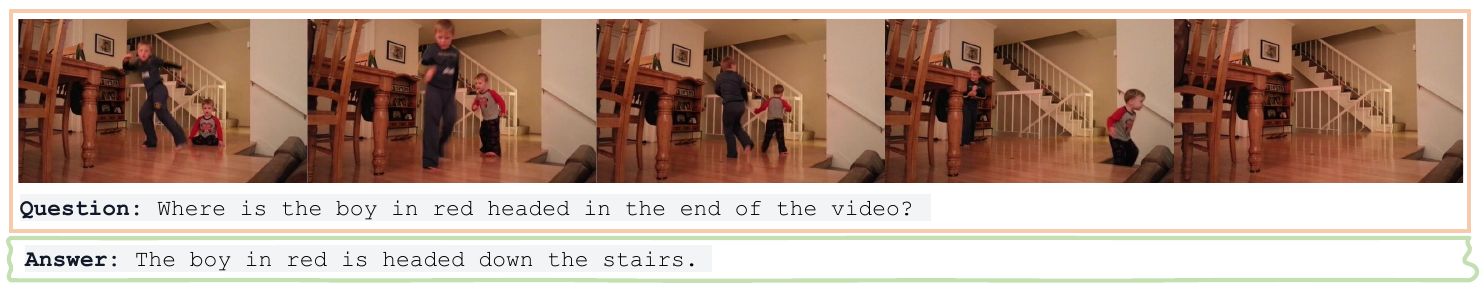}
    \caption{Showcases of the \lexvi~'s performance on verbatim text generation, sentence retrieval, multi-image reasoning, and video question answering, demonstrating its long-context understanding.
    }
    \label{fig:seeker_showcase}
\end{figure*}

Figure~\ref{fig:seeker_showcase} showcases the \lexvi~ model's performance on three tasks, emphasizing its long-context capabilities.
In the verbatim generation task, \lexvi~ read text from the arXiv paper, indicating its coherent narratives given extended multimodal context.
For the first sentence retrieval task, it efficiently navigated and extracted key sentences from extensive texts without utilizing the OCR model.
In the task of reasoning across multiple images, the model effectively grounds the text in the specific image as required.
At the bottom of Figure~\ref{fig:seeker_showcase}, we observe that \lexvi~ can also generalize to multi-frame video understanding.
We compare \lexvis~ with DeepSeek-VL-7B on identifying the document titles in Table~\ref{tab:seeker_comparison}. \lexvi~ excels at capturing character-level details. These results illustrate \lexvi~'s proficiency in handling long-context multimodal tasks, marking a significant advancement in \mllms~.

\begin{table}[t!]
\caption{Comparisons of MLLMs' Instruction-Following Character-Level Recognition.}
\label{tab:seeker_comparison}
\centering
\begin{tcolorbox}[colback=splashedwhite,boxsep=1pt,left=2pt,right=5pt,top=5pt,bottom=3pt,boxrule=0.5pt]
\centering
\small
\begin{tabular}{p{0.6\columnwidth} p{0.3\columnwidth}}
\VarSty{ {\bf {\textit{Image}}} } & \VarSty{{\bf {\textit{Text Prompt}}}}: What is the title of the document? \\ 
\\
\vspace{-10mm}
\multirow{4}{*}{
\includegraphics[height=5cm]{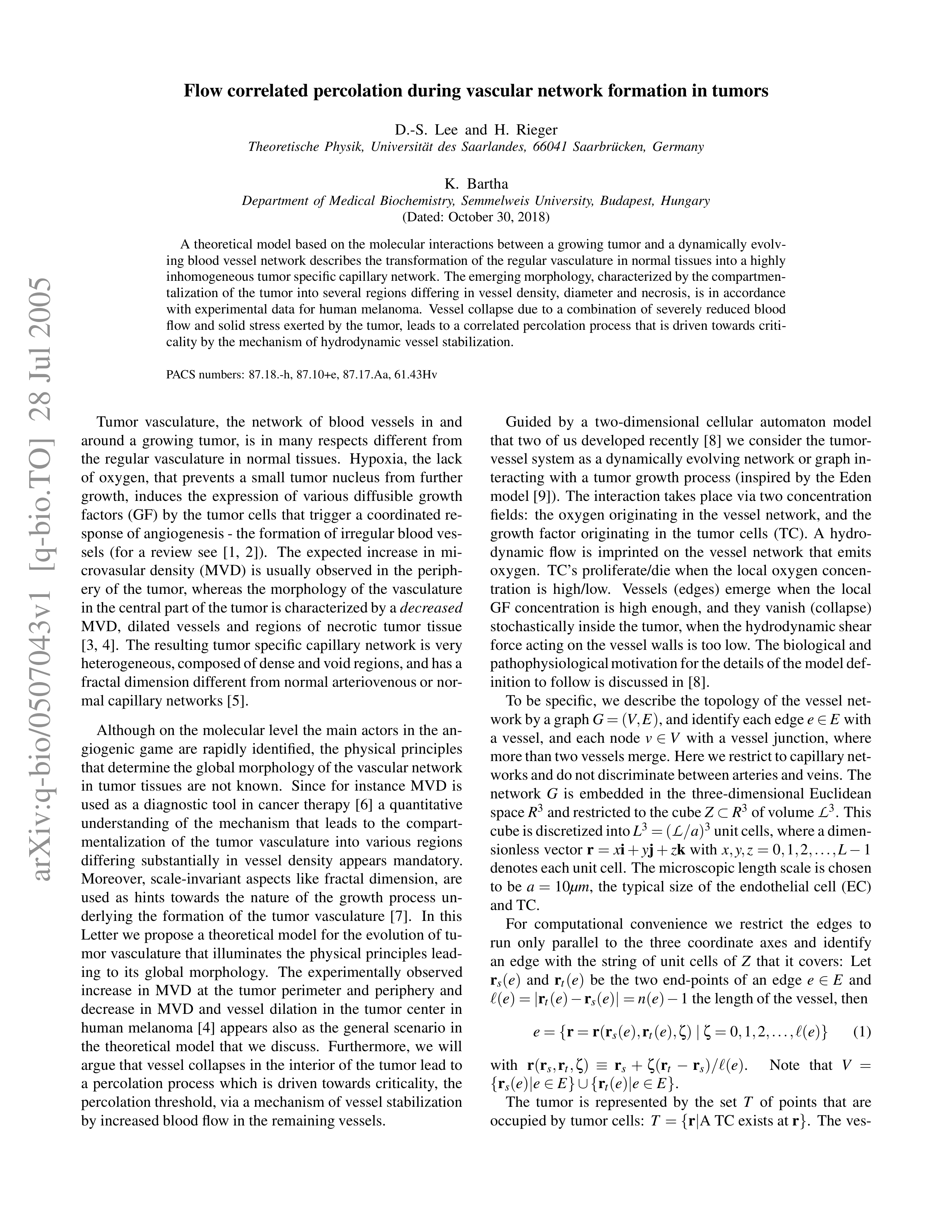} 
\includegraphics[height=5cm]{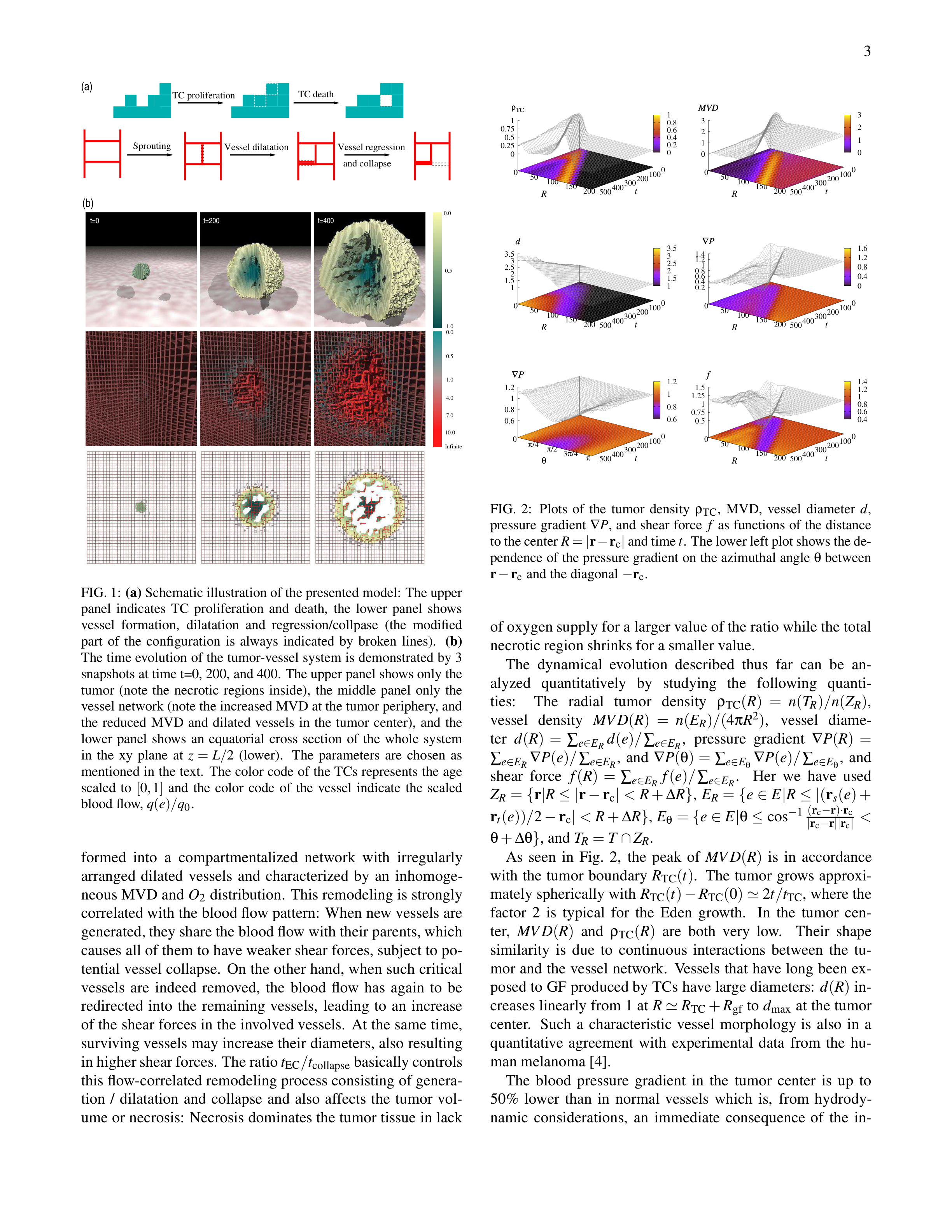} 
}
& \VarSty{ {\bf {\textit{Reference}}: }} Flow correlated percolation \textcolor{persiangreen}{during} vascular network formation in tumors  \\
\vspace{6mm}
\\
& \VarSty{ {\bf DeepSeek-VL-7B:}} Flow correlated percolation \textcolor{cadmiumred}{driving} vascular network formation in tumors. \\ 
\\
& \VarSty{ {\bf \lexvi~: }} Flow correlated percolation \textcolor{persiangreen}{during} vascular network formation in tumors  \\
\vspace{2.5cm}
\end{tabular}
\end{tcolorbox}
\end{table}

\section{Conclusion}
In this paper, we present \lexvi~, which advances the field of long-context comprehension in multimodal large language models. By enhancing the processing of lengthy texts presented in visual formats and continual instruction-tuning on extended context tasks, \lexvi~ surpasses existing multimodal large language models in handling extensive multimodal contexts. Our \lexvi~ also shows efficiency compared with OCR-based approach in terms of better long context extrapolation and inference efficiency.
Additionally, it generalizes effectively across various domains, including video question answering. We hope our work paves the way for future studies in efficiently handling long multimodal contexts.

\begin{ack}
This research was supported by the ICB cooperative agreement W911NF-19-2-0026. The writers’ opinions and conclusions in this publication are their own and should not be construed as representing the sponsors.
\end{ack}


\medskip

{
\small
\bibliographystyle{plain}
\bibliography{reference}

\begin{thebibliography}{10}

\bibitem{bai2023qwen}
Jinze Bai, Shuai Bai, Shusheng Yang, Shijie Wang, Sinan Tan, Peng Wang, Junyang Lin, Chang Zhou, and Jingren Zhou.
\newblock Qwen-vl: A frontier large vision-language model with versatile abilities.
\newblock {\em arXiv preprint arXiv:2308.12966}, 2023.

\bibitem{bai2023qwenvl}
Jinze Bai, Shuai Bai, Shusheng Yang, Shijie Wang, Sinan Tan, Peng Wang, Junyang Lin, Chang Zhou, and Jingren Zhou.
\newblock Qwen-vl: A versatile vision-language model for understanding, localization, text reading, and beyond, 2023.

\bibitem{blecher2023nougat}
Lukas Blecher, Guillem Cucurull, Thomas Scialom, and Robert Stojnic.
\newblock Nougat: Neural optical understanding for academic documents, 2023.

\bibitem{chen2024longlora}
Yukang Chen, Shengju Qian, Haotian Tang, Xin Lai, Zhijian Liu, Song Han, and Jiaya Jia.
\newblock Longlora: Efficient fine-tuning of long-context large language models, 2024.

\bibitem{vicuna2023}
Wei-Lin Chiang, Zhuohan Li, Zi~Lin, Ying Sheng, Zhanghao Wu, Hao Zhang, Lianmin Zheng, Siyuan Zhuang, Yonghao Zhuang, Joseph~E. Gonzalez, Ion Stoica, and Eric~P. Xing.
\newblock Vicuna: An open-source chatbot impressing gpt-4 with 90\%* chatgpt quality, March 2023.

\bibitem{2023opencompass}
OpenCompass Contributors.
\newblock Opencompass: A universal evaluation platform for foundation models.
\newblock \url{https://github.com/open-compass/opencompass}, 2023.

\bibitem{dai2023instructblip}
Wenliang Dai, Junnan Li, Dongxu Li, Anthony Meng~Huat Tiong, Junqi Zhao, Weisheng Wang, Boyang Li, Pascale Fung, and Steven Hoi.
\newblock Instructblip: Towards general-purpose vision-language models with instruction tuning.
\newblock {\em arXiv preprint arXiv:2305.06500}, 2023.

\bibitem{deepseekai2024deepseek}
DeepSeek-AI, :, Xiao Bi, Deli Chen, Guanting Chen, Shanhuang Chen, Damai Dai, Chengqi Deng, Honghui Ding, Kai Dong, Qiushi Du, Zhe Fu, Huazuo Gao, Kaige Gao, Wenjun Gao, Ruiqi Ge, Kang Guan, Daya Guo, Jianzhong Guo, Guangbo Hao, Zhewen Hao, Ying He, Wenjie Hu, Panpan Huang, Erhang Li, Guowei Li, Jiashi Li, Yao Li, Y.~K. Li, Wenfeng Liang, Fangyun Lin, A.~X. Liu, Bo~Liu, Wen Liu, Xiaodong Liu, Xin Liu, Yiyuan Liu, Haoyu Lu, Shanghao Lu, Fuli Luo, Shirong Ma, Xiaotao Nie, Tian Pei, Yishi Piao, Junjie Qiu, Hui Qu, Tongzheng Ren, Zehui Ren, Chong Ruan, Zhangli Sha, Zhihong Shao, Junxiao Song, Xuecheng Su, Jingxiang Sun, Yaofeng Sun, Minghui Tang, Bingxuan Wang, Peiyi Wang, Shiyu Wang, Yaohui Wang, Yongji Wang, Tong Wu, Y.~Wu, Xin Xie, Zhenda Xie, Ziwei Xie, Yiliang Xiong, Hanwei Xu, R.~X. Xu, Yanhong Xu, Dejian Yang, Yuxiang You, Shuiping Yu, Xingkai Yu, B.~Zhang, Haowei Zhang, Lecong Zhang, Liyue Zhang, Mingchuan Zhang, Minghua Zhang, Wentao Zhang, Yichao Zhang, Chenggang Zhao, Yao Zhao, Shangyan Zhou, Shunfeng
  Zhou, Qihao Zhu, and Yuheng Zou.
\newblock Deepseek llm: Scaling open-source language models with longtermism, 2024.

\bibitem{dosovitskiy2021image}
Alexey Dosovitskiy, Lucas Beyer, Alexander Kolesnikov, Dirk Weissenborn, Xiaohua Zhai, Thomas Unterthiner, Mostafa Dehghani, Matthias Minderer, Georg Heigold, Sylvain Gelly, Jakob Uszkoreit, and Neil Houlsby.
\newblock An image is worth 16x16 words: Transformers for image recognition at scale, 2021.

\bibitem{fu2024blink}
Xingyu Fu, Yushi Hu, Bangzheng Li, Yu~Feng, Haoyu Wang, Xudong Lin, Dan Roth, Noah~A. Smith, Wei-Chiu Ma, and Ranjay Krishna.
\newblock Blink: Multimodal large language models can see but not perceive, 2024.

\bibitem{gao2024improving}
Tianyu Gao, Zirui Wang, Adithya Bhaskar, and Danqi Chen.
\newblock Improving language understanding from screenshots, 2024.

\bibitem{hu2021lora}
Edward~J. Hu, Yelong Shen, Phillip Wallis, Zeyuan Allen-Zhu, Yuanzhi Li, Shean Wang, Lu~Wang, and Weizhu Chen.
\newblock Lora: Low-rank adaptation of large language models, 2021.

\bibitem{hu2024minicpm}
Shengding Hu, Yuge Tu, Xu~Han, Chaoqun He, Ganqu Cui, Xiang Long, Zhi Zheng, Yewei Fang, Yuxiang Huang, Weilin Zhao, Xinrong Zhang, Zheng~Leng Thai, Kaihuo Zhang, Chongyi Wang, Yuan Yao, Chenyang Zhao, Jie Zhou, Jie Cai, Zhongwu Zhai, Ning Ding, Chao Jia, Guoyang Zeng, Dahai Li, Zhiyuan Liu, and Maosong Sun.
\newblock Minicpm: Unveiling the potential of small language models with scalable training strategies, 2024.

\bibitem{jiang2024mantis}
Dongfu Jiang, Xuan He, Huaye Zeng, Cong Wei, Max Ku, Qian Liu, and Wenhu Chen.
\newblock Mantis: Interleaved multi-image instruction tuning, 2024.

\bibitem{jin2024llm}
Hongye Jin, Xiaotian Han, Jingfeng Yang, Zhimeng Jiang, Zirui Liu, Chia-Yuan Chang, Huiyuan Chen, and Xia Hu.
\newblock Llm maybe longlm: Self-extend llm context window without tuning, 2024.

\bibitem{Kembhavi2016ADI}
Aniruddha Kembhavi, Michael Salvato, Eric Kolve, Minjoon Seo, Hannaneh Hajishirzi, and Ali Farhadi.
\newblock A diagram is worth a dozen images.
\newblock {\em ArXiv}, abs/1603.07396, 2016.

\bibitem{kirillov2023segment}
Alexander Kirillov, Eric Mintun, Nikhila Ravi, Hanzi Mao, Chloe Rolland, Laura Gustafson, Tete Xiao, Spencer Whitehead, Alexander~C. Berg, Wan-Yen Lo, Piotr Dollár, and Ross Girshick.
\newblock Segment anything, 2023.

\bibitem{laurençon2024matters}
Hugo Laurençon, Léo Tronchon, Matthieu Cord, and Victor Sanh.
\newblock What matters when building vision-language models?, 2024.

\bibitem{li2023seedbench}
Bohao Li, Rui Wang, Guangzhi Wang, Yuying Ge, Yixiao Ge, and Ying Shan.
\newblock Seed-bench: Benchmarking multimodal llms with generative comprehension, 2023.

\bibitem{li2023monkey}
Zhang Li, Biao Yang, Qiang Liu, Zhiyin Ma, Shuo Zhang, Jingxu Yang, Yabo Sun, Yuliang Liu, and Xiang Bai.
\newblock Monkey: Image resolution and text label are important things for large multi-modal models.
\newblock {\em arXiv preprint arXiv:2311.06607}, 2023.

\bibitem{lin-2004-rouge}
Chin-Yew Lin.
\newblock {ROUGE}: A package for automatic evaluation of summaries.
\newblock In {\em Text Summarization Branches Out}, pages 74--81, Barcelona, Spain, July 2004. Association for Computational Linguistics.

\bibitem{lin2015microsoft}
Tsung-Yi Lin, Michael Maire, Serge Belongie, Lubomir Bourdev, Ross Girshick, James Hays, Pietro Perona, Deva Ramanan, C.~Lawrence Zitnick, and Piotr Dollár.
\newblock Microsoft coco: Common objects in context, 2015.

\bibitem{liu2023improved}
Haotian Liu, Chunyuan Li, Yuheng Li, and Yong~Jae Lee.
\newblock Improved baselines with visual instruction tuning.
\newblock {\em arXiv preprint arXiv:2310.03744}, 2023.

\bibitem{liu2023improvedllava}
Haotian Liu, Chunyuan Li, Yuheng Li, and Yong~Jae Lee.
\newblock Improved baselines with visual instruction tuning, 2023.

\bibitem{liu2024llavanext}
Haotian Liu, Chunyuan Li, Yuheng Li, Bo~Li, Yuanhan Zhang, Sheng Shen, and Yong~Jae Lee.
\newblock Llava-next: Improved reasoning, ocr, and world knowledge, January 2024.

\bibitem{liu2023llava}
Haotian Liu, Chunyuan Li, Qingyang Wu, and Yong~Jae Lee.
\newblock Visual instruction tuning.
\newblock In {\em NeurIPS}, 2023.

\bibitem{liu2023visual}
Haotian Liu, Chunyuan Li, Qingyang Wu, and Yong~Jae Lee.
\newblock Visual instruction tuning.
\newblock {\em arXiv preprint arXiv:2304.08485}, 2023.

\bibitem{liu2023lost}
Nelson~F. Liu, Kevin Lin, John Hewitt, Ashwin Paranjape, Michele Bevilacqua, Fabio Petroni, and Percy Liang.
\newblock Lost in the middle: How language models use long contexts, 2023.

\bibitem{liu2024mmbench}
Yuan Liu, Haodong Duan, Yuanhan Zhang, Bo~Li, Songyang Zhang, Wangbo Zhao, Yike Yuan, Jiaqi Wang, Conghui He, Ziwei Liu, Kai Chen, and Dahua Lin.
\newblock Mmbench: Is your multi-modal model an all-around player?, 2024.

\bibitem{loshchilov2019decoupled}
Ilya Loshchilov and Frank Hutter.
\newblock Decoupled weight decay regularization, 2019.

\bibitem{lu2024deepseekvl}
Haoyu Lu, Wen Liu, Bo~Zhang, Bingxuan Wang, Kai Dong, Bo~Liu, Jingxiang Sun, Tongzheng Ren, Zhuoshu Li, Hao Yang, Yaofeng Sun, Chengqi Deng, Hanwei Xu, Zhenda Xie, and Chong Ruan.
\newblock Deepseek-vl: Towards real-world vision-language understanding, 2024.

\bibitem{lu2023vim}
Yujie Lu, Xiujun Li, William~Yang Wang, and Yejin Choi.
\newblock Vim: Probing multimodal large language models for visual embedded instruction following, 2023.

\bibitem{masry2022chartqa}
Ahmed Masry, Do~Xuan Long, Jia~Qing Tan, Shafiq Joty, and Enamul Hoque.
\newblock Chartqa: A benchmark for question answering about charts with visual and logical reasoning, 2022.

\bibitem{mckinzie2024mm1}
Brandon McKinzie, Zhe Gan, Jean-Philippe Fauconnier, Sam Dodge, Bowen Zhang, Philipp Dufter, Dhruti Shah, Xianzhi Du, Futang Peng, Floris Weers, Anton Belyi, Haotian Zhang, Karanjeet Singh, Doug Kang, Ankur Jain, Hongyu Hè, Max Schwarzer, Tom Gunter, Xiang Kong, Aonan Zhang, Jianyu Wang, Chong Wang, Nan Du, Tao Lei, Sam Wiseman, Guoli Yin, Mark Lee, Zirui Wang, Ruoming Pang, Peter Grasch, Alexander Toshev, and Yinfei Yang.
\newblock Mm1: Methods, analysis \& insights from multimodal llm pre-training, 2024.

\bibitem{chatgpt2022}
OpenAI.
\newblock Chatgpt.
\newblock {\em https://chat.openai.com/}, 2022.

\bibitem{gpt4}
OpenAI.
\newblock Gpt-4: Technical report.
\newblock {\em arXiv preprint arXiv:2303.08774}, 2023.

\bibitem{gpt4-v}
OpenAI.
\newblock Gpt-4v(ision) system card.
\newblock {\em https://openai.com/research/gpt-4v-system-card}, 2023.

\bibitem{gpt-4o}
OpenAI.
\newblock Gpt-4o.
\newblock {\em https://openai.com/index/hello-gpt-4o}, 2024.

\bibitem{rust-etal-2023-pixel}
Phillip Rust, Jonas~F. Lotz, Emanuele Bugliarello, Elizabeth Salesky, Miryam de~Lhoneux, and Desmond Elliott.
\newblock Language modelling with pixels.
\newblock In {\em The Eleventh International Conference on Learning Representations}, 2023.

\bibitem{singh2019vqa}
Amanpreet Singh, Vivek Natarajan, Meet Shah, Yu~Jiang, Xinlei Chen, Dhruv Batra, Devi Parikh, and Marcus Rohrbach.
\newblock Towards vqa models that can read, 2019.

\bibitem{suhr-etal-2019-corpus}
Alane Suhr, Stephanie Zhou, Ally Zhang, Iris Zhang, Huajun Bai, and Yoav Artzi.
\newblock A corpus for reasoning about natural language grounded in photographs.
\newblock In Anna Korhonen, David Traum, and Llu{\'\i}s M{\`a}rquez, editors, {\em Proceedings of the 57th Annual Meeting of the Association for Computational Linguistics}, pages 6418--6428, Florence, Italy, July 2019. Association for Computational Linguistics.

\bibitem{team2023gemini}
Gemini Team, Rohan Anil, Sebastian Borgeaud, Yonghui Wu, Jean-Baptiste Alayrac, Jiahui Yu, Radu Soricut, Johan Schalkwyk, Andrew~M Dai, Anja Hauth, et~al.
\newblock Gemini: a family of highly capable multimodal models.
\newblock {\em arXiv preprint arXiv:2312.11805}, 2023.

\bibitem{touvron2023llama}
Hugo Touvron, Thibaut Lavril, Gautier Izacard, Xavier Martinet, Marie-Anne Lachaux, Timoth{\'e}e Lacroix, Baptiste Rozi{\`e}re, Naman Goyal, Eric Hambro, Faisal Azhar, et~al.
\newblock Llama: Open and efficient foundation language models.
\newblock {\em arXiv preprint arXiv:2302.13971}, 2023.

\bibitem{touvron2023llama2}
Hugo Touvron, Louis Martin, Kevin Stone, Peter Albert, Amjad Almahairi, Yasmine Babaei, Nikolay Bashlykov, Soumya Batra, Prajjwal Bhargava, Shruti Bhosale, et~al.
\newblock Llama 2: Open foundation and fine-tuned chat models.
\newblock {\em arXiv preprint arXiv:2307.09288}, 2023.

\bibitem{tworkowski2023focused}
Szymon Tworkowski, Konrad Staniszewski, Mikołaj Pacek, Yuhuai Wu, Henryk Michalewski, and Piotr Miłoś.
\newblock Focused transformer: Contrastive training for context scaling, 2023.

\bibitem{wang2021images}
Yulin Wang, Rui Huang, Shiji Song, Zeyi Huang, and Gao Huang.
\newblock Not all images are worth 16x16 words: Dynamic transformers for efficient image recognition, 2021.

\bibitem{wu2024comprehensive}
Tianhe Wu, Kede Ma, Jie Liang, Yujiu Yang, and Lei Zhang.
\newblock A comprehensive study of multimodal large language models for image quality assessment, 2024.

\bibitem{xiong2023effective}
Wenhan Xiong, Jingyu Liu, Igor Molybog, Hejia Zhang, Prajjwal Bhargava, Rui Hou, Louis Martin, Rashi Rungta, Karthik~Abinav Sankararaman, Barlas Oguz, Madian Khabsa, Han Fang, Yashar Mehdad, Sharan Narang, Kshitiz Malik, Angela Fan, Shruti Bhosale, Sergey Edunov, Mike Lewis, Sinong Wang, and Hao Ma.
\newblock Effective long-context scaling of foundation models, 2023.

\bibitem{yu2023mmvet}
Weihao Yu, Zhengyuan Yang, Linjie Li, Jianfeng Wang, Kevin Lin, Zicheng Liu, Xinchao Wang, and Lijuan Wang.
\newblock Mm-vet: Evaluating large multimodal models for integrated capabilities.
\newblock {\em arXiv preprint arXiv:2308.02490}, 2023.

\bibitem{zhai2023sigmoid}
Xiaohua Zhai, Basil Mustafa, Alexander Kolesnikov, and Lucas Beyer.
\newblock Sigmoid loss for language image pre-training, 2023.

\bibitem{zong2022selfslimmed}
Zhuofan Zong, Kunchang Li, Guanglu Song, Yali Wang, Yu~Qiao, Biao Leng, and Yu~Liu.
\newblock Self-slimmed vision transformer, 2022.

\end{thebibliography}
}

\clearpage

\appendix
\addcontentsline{toc}{section}{Appendix} 
\part{Appendix} 
\parttoc 

\clearpage
\section{Implementation Details of \lexvi~}
\subsection{Training Loss Curve}
\begin{figure*}[t!]
    \centering
    \minipage{.49\linewidth}
    \includegraphics[width=\linewidth]{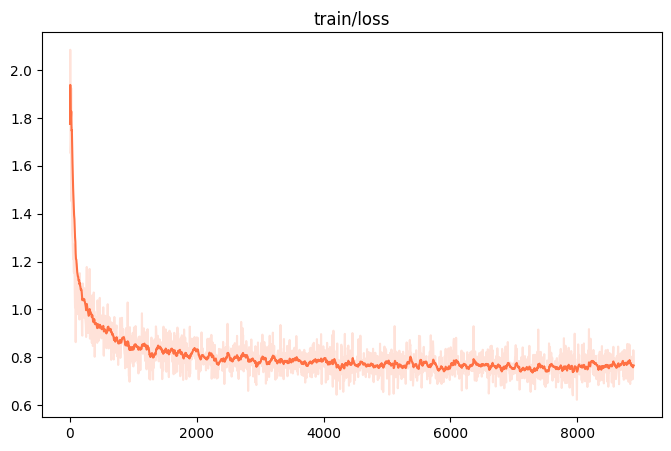}
    \subcaption{\lexvi~}
    \endminipage\hfill
    \minipage{.49\linewidth}
    \includegraphics[width=\linewidth]{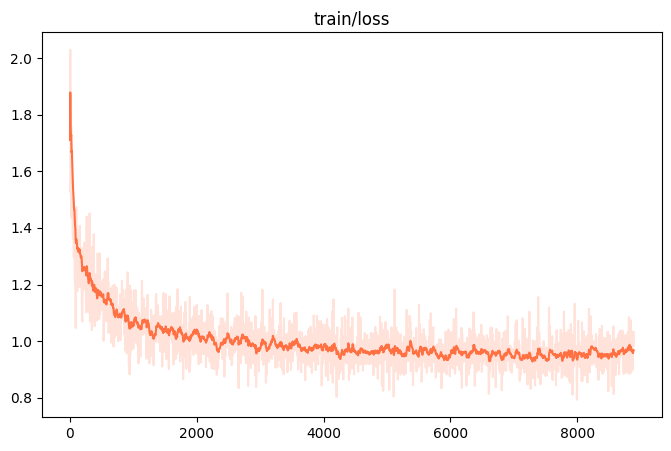}
    \subcaption{\lexvit~}
    \endminipage\hfill
    \caption{Training Loss Curve.}
    \label{fig:training_curve}
\end{figure*}

In Figure~\ref{fig:training_curve}, we show the training loss curve of our \lexvi~ and \lexvit~. Though both model have a quick loss drop initially, we observe a smoother and more consistent decrease of \lexvi~ than \lexvit~. In the end, \lexvi~ stabilizes at a lower loss value, suggesting its potentially better generalization capabilities than \lexvit~.

\subsection{Evaluation Benchmarks and Metrics}
\label{sec:evaluation_details}
We consider four long-form multi-image input tasks: 1) \texttt{Index}: the multiple-choice image indexing task, given a sequence of images and a question, the model selects the option with the index of the image that contains the answer, 2) \texttt{SentRetrie}: the sentence retrieval task, given a sequence of images of rendered text sampled from Wikipedia, the model is required to retrieve the first sentence from the first image, 3) \texttt{ArxivQA}: the question answering on arxiv documents, the model is required to answer the question according to visual image of arxiv documents. 4) \texttt{PassKey}: the passkey retrieval task slightly modified for multimodal model, given the sentence with a masked word, the model need to answer what is the masked word by reading the visually-situated text content from arxiv document.
We consider two long-form text output tasks: 1) \texttt{ArxivVerb}: extract text from the image of arxiv documents verbatim, 2) \texttt{WikiVerb}: extract text from the image of rendered text from Wikipedia verbatim.

\clearpage
\section{More Analysis}
\subsection{Tradeoff of Compact Context Length and High Resolution}
In Figure~\ref{fig:resolution_or_compact}, we show GPT-4-Vision with low and high resolution setting on first-sentence-retrieval. With high-resolution mode, more tokens will be used to represent the same image. Although high-resolution usually brings more details and better performance, we can see it tradeoffs capability of extrapolating long page document understanding. And thus only GPT-4-Vision low-resolution model preserves the performance in this probing task. On the right we can see that high-resolution usually take more image tokens to represent text-rich image than text tokens of OCR-extracted content, and thus even drops more quickly than feeding text.
\begin{center}
    \includegraphics[width=0.32\textwidth]{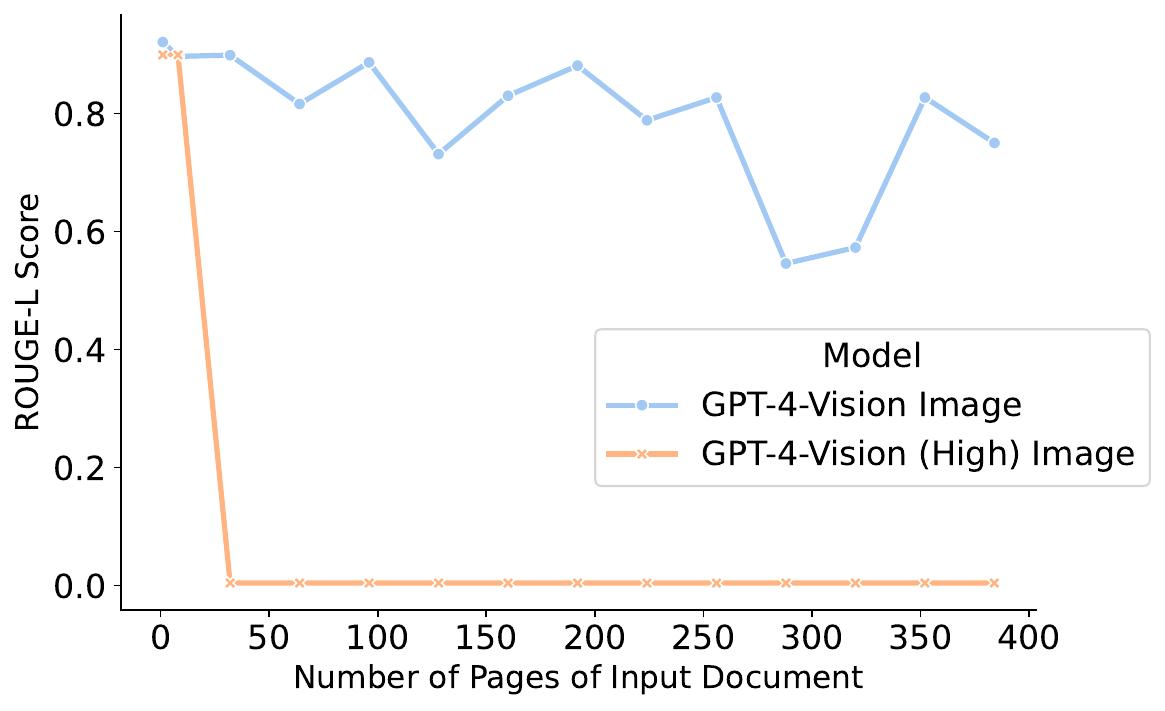}
    \includegraphics[width=0.32\textwidth]{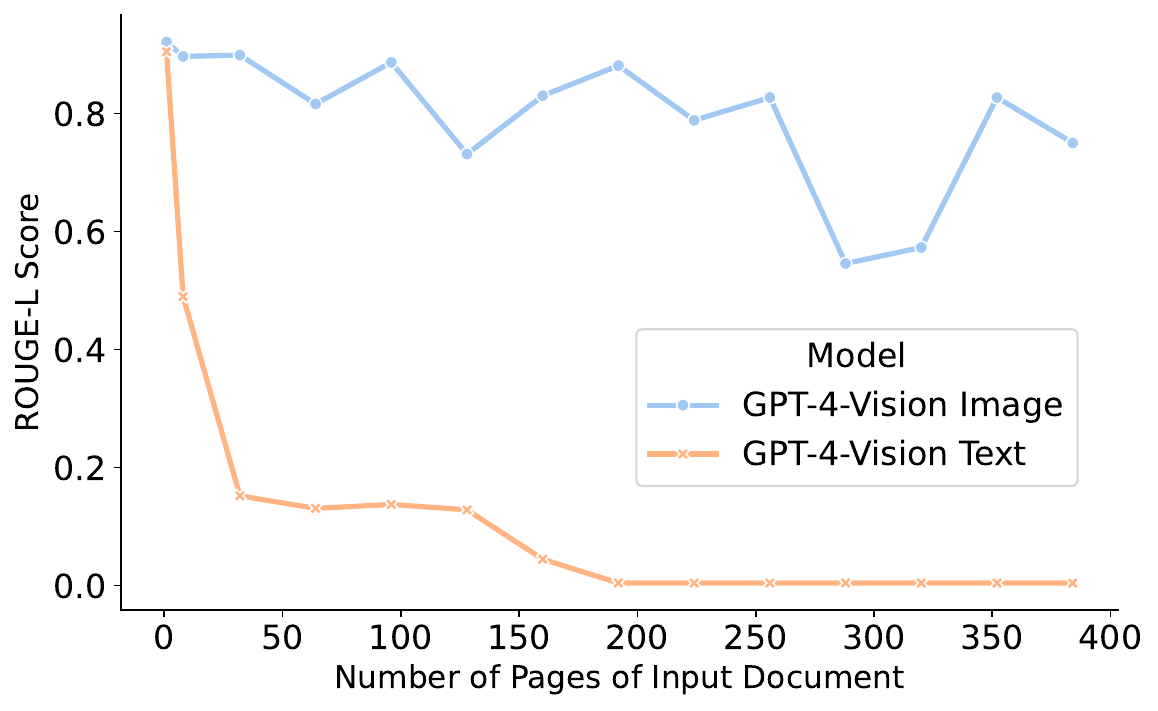}
    \includegraphics[width=0.32\textwidth]{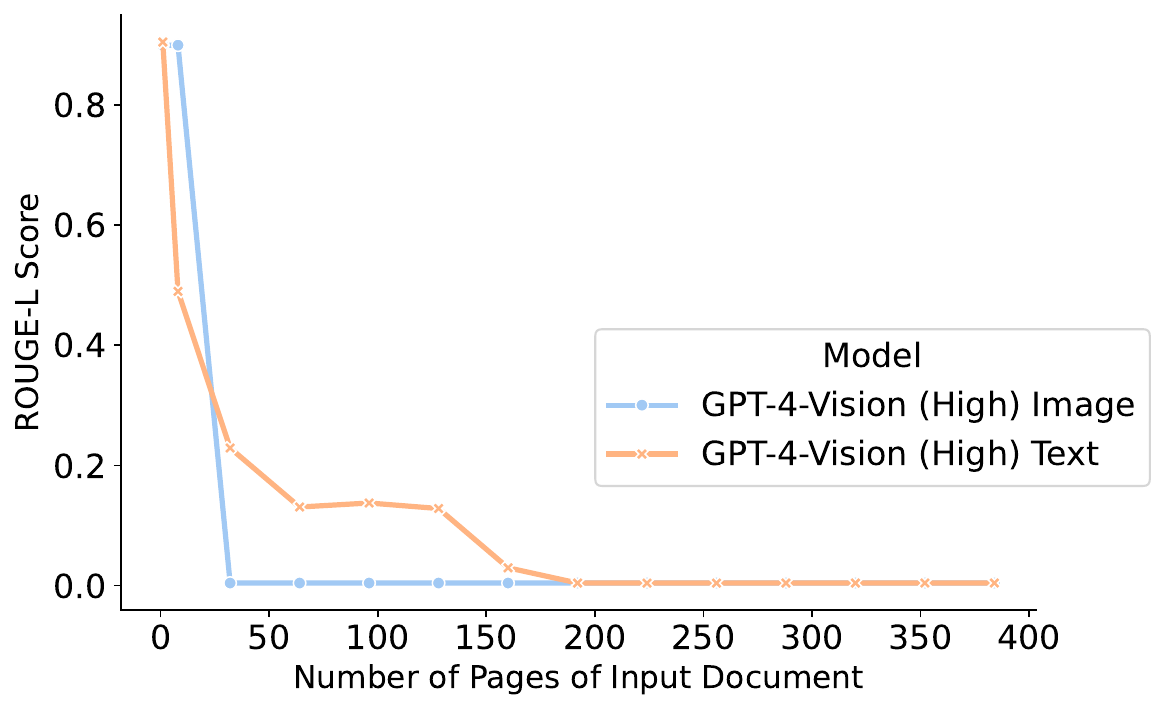}
    \captionof{figure}{Performance plot on First-Sentence-Retrieval task.
    }
    \label{fig:resolution_or_compact}
\end{center}%

\clearpage
\section{Long-Context Multimodal Tasks}
\subsection{Task Examples}
\begin{figure*}[t!]
    \centering
    \includegraphics[width=\linewidth]{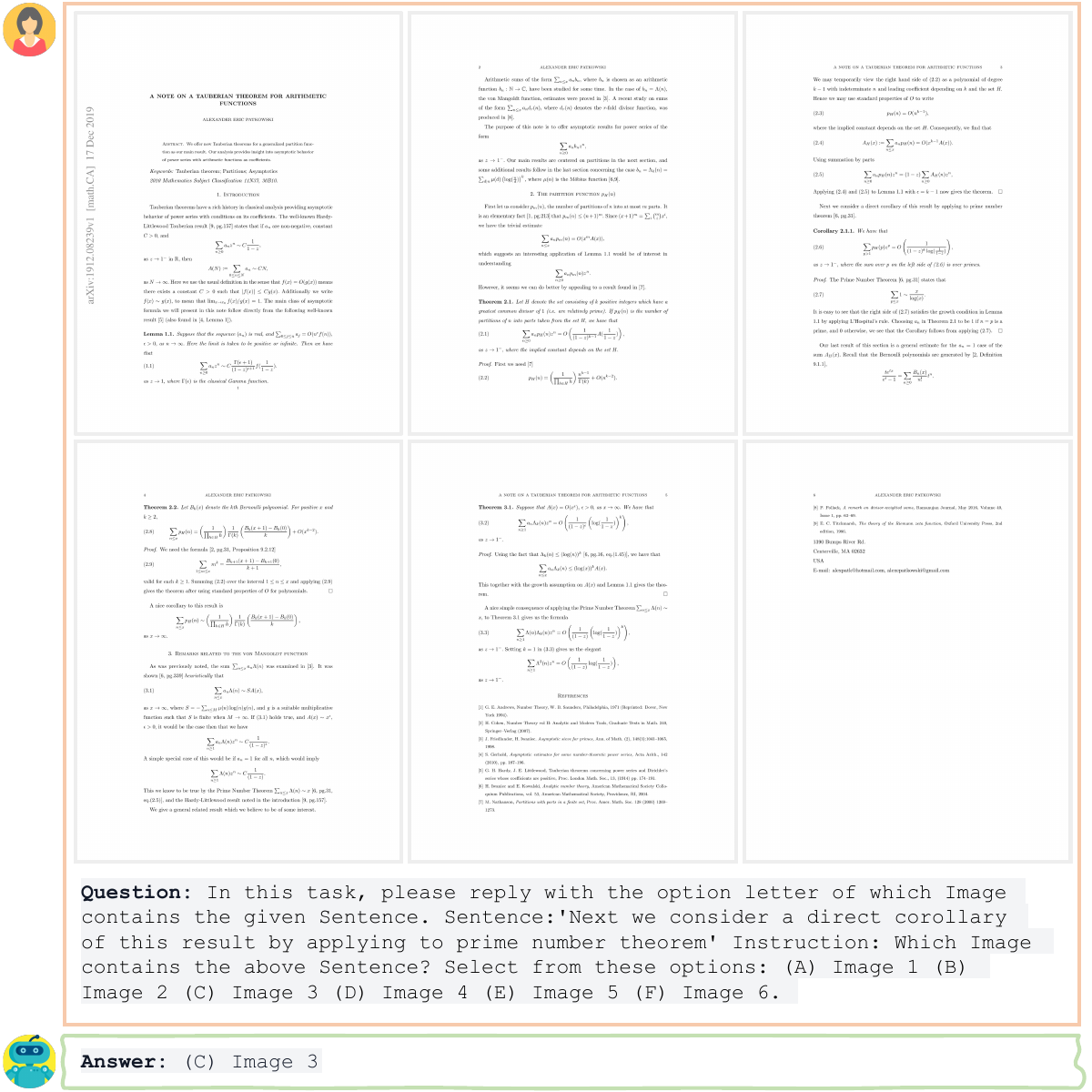}
    \caption{Task \texttt{Index}.
    }
    \label{fig:task_index}
\end{figure*}

\begin{figure*}[t!]
    \centering
    \includegraphics[width=\linewidth]{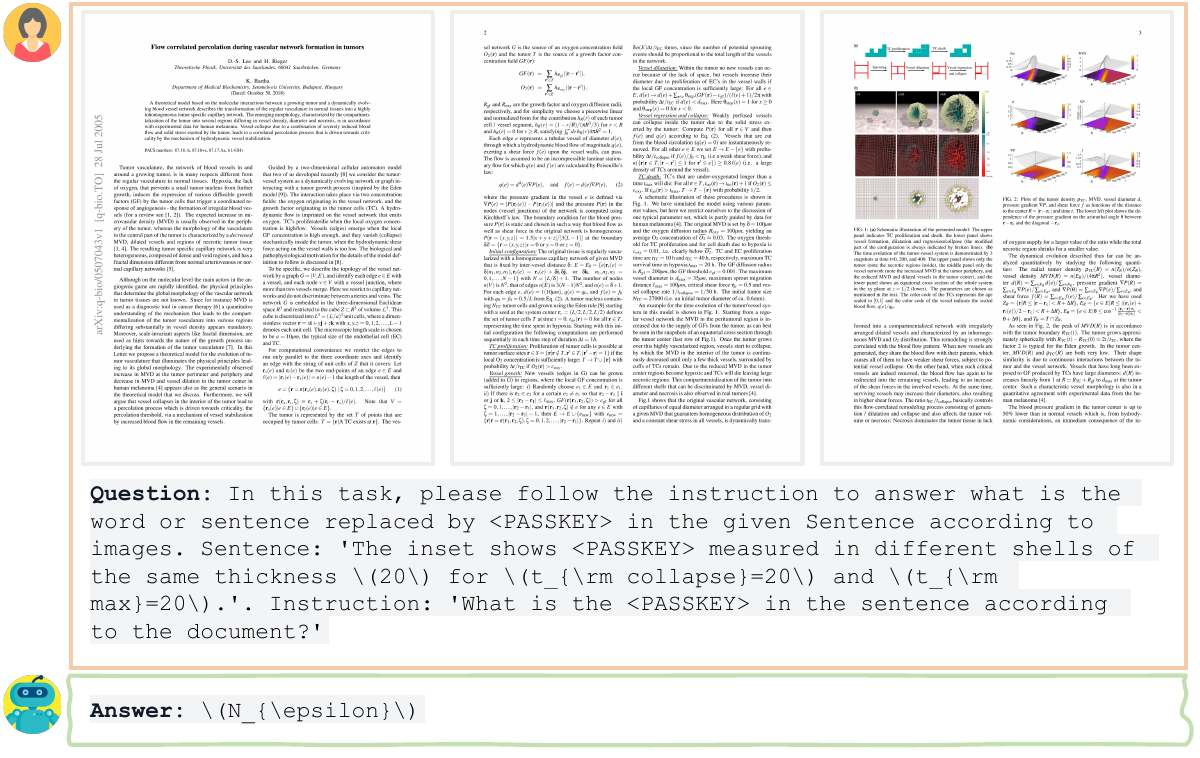}
    \caption{Task \texttt{PassKey}.
    }
    \label{fig:task_passkey}
\end{figure*}

\begin{figure*}[t!]
    \centering
    \includegraphics[width=\linewidth]{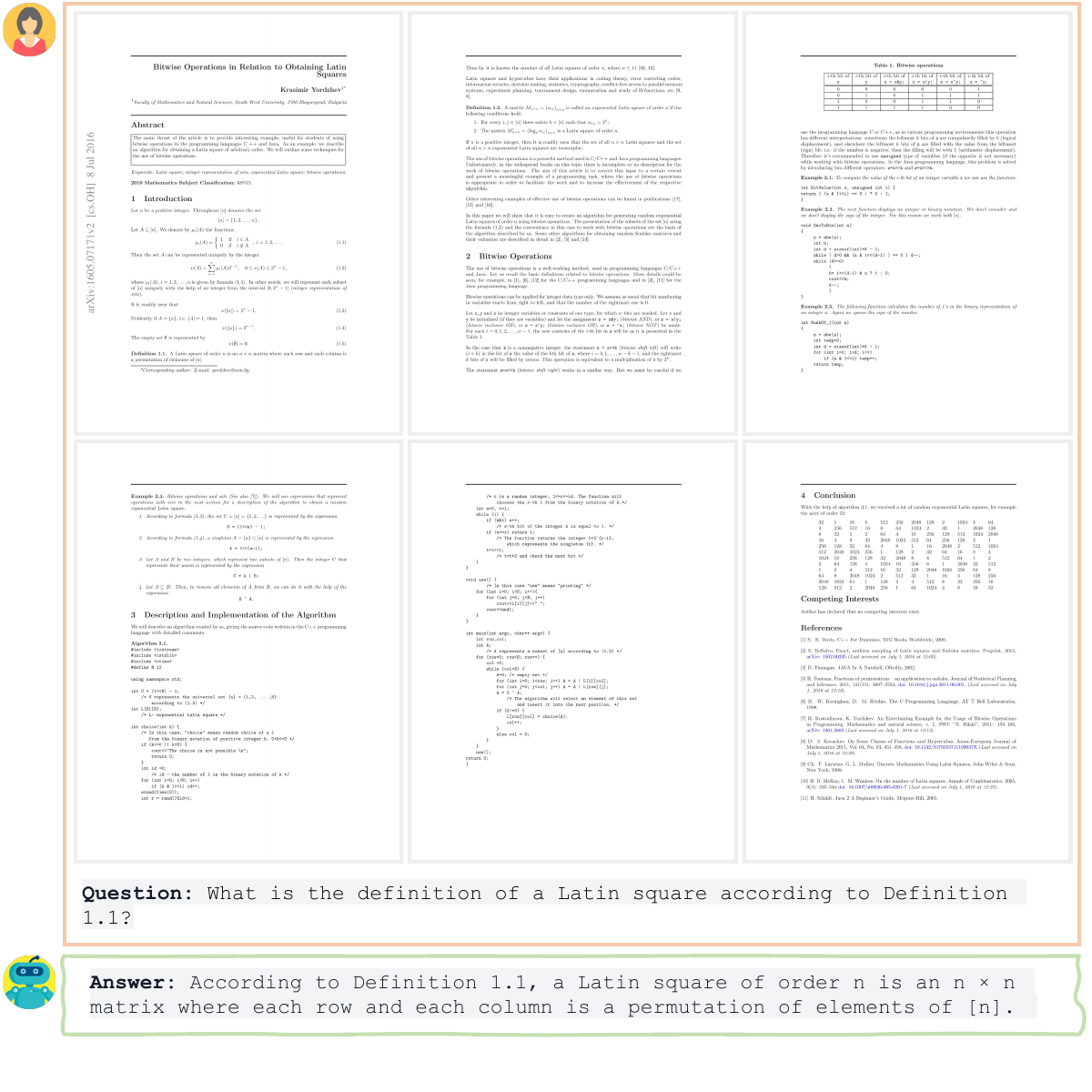}
    \caption{Task \texttt{ArxivQA}.
    }
    \label{fig:task_arxivqa}
\end{figure*}

\begin{figure*}[t!]
    \centering
    \includegraphics[width=\linewidth]{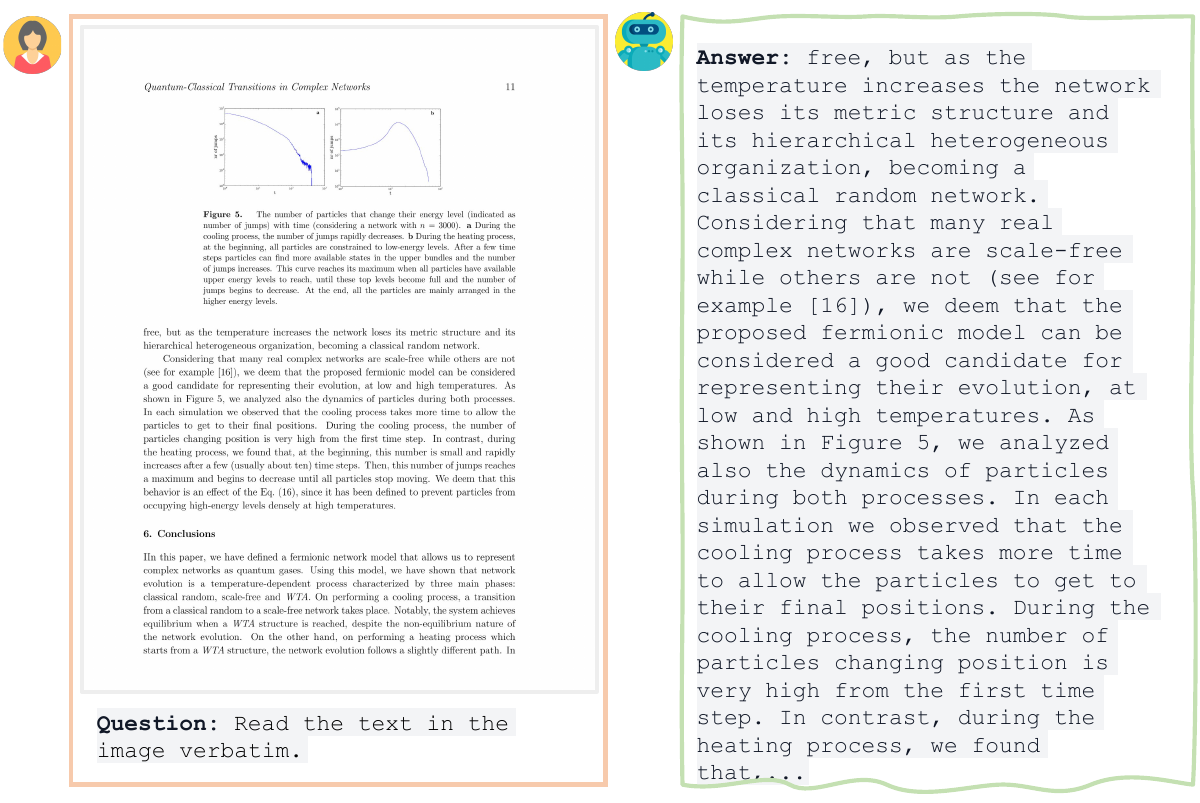}
    \caption{Task \texttt{ArxivVerbatim}.
    }
    \label{fig:task_arxivverbatim}
\end{figure*}

\begin{figure*}[t!]
    \centering
    \includegraphics[width=\linewidth]{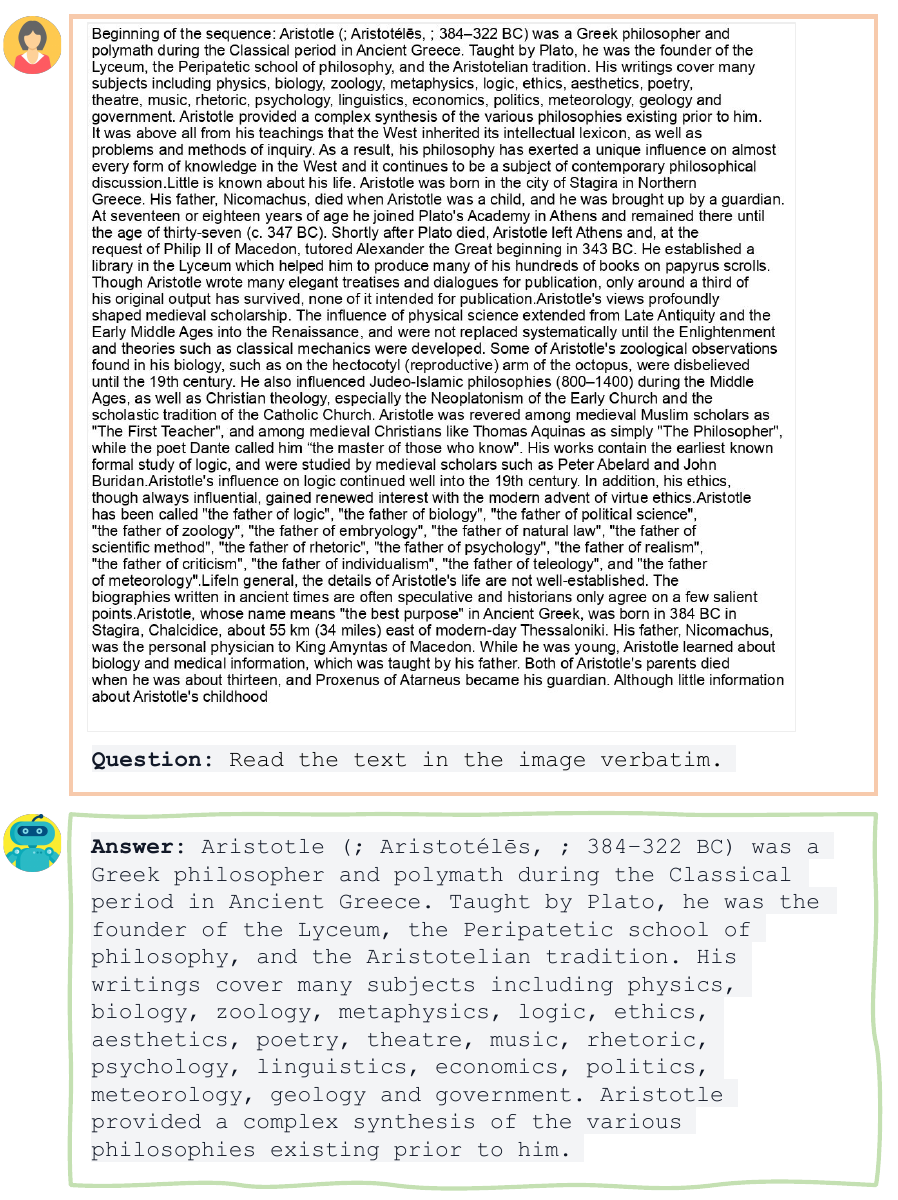}
    \caption{Task \texttt{WikiVerbatim}.
    }
    \label{fig:task_wikiverbatim}
\end{figure*}

\begin{figure*}[t!]
    \centering
    \includegraphics[width=\linewidth]{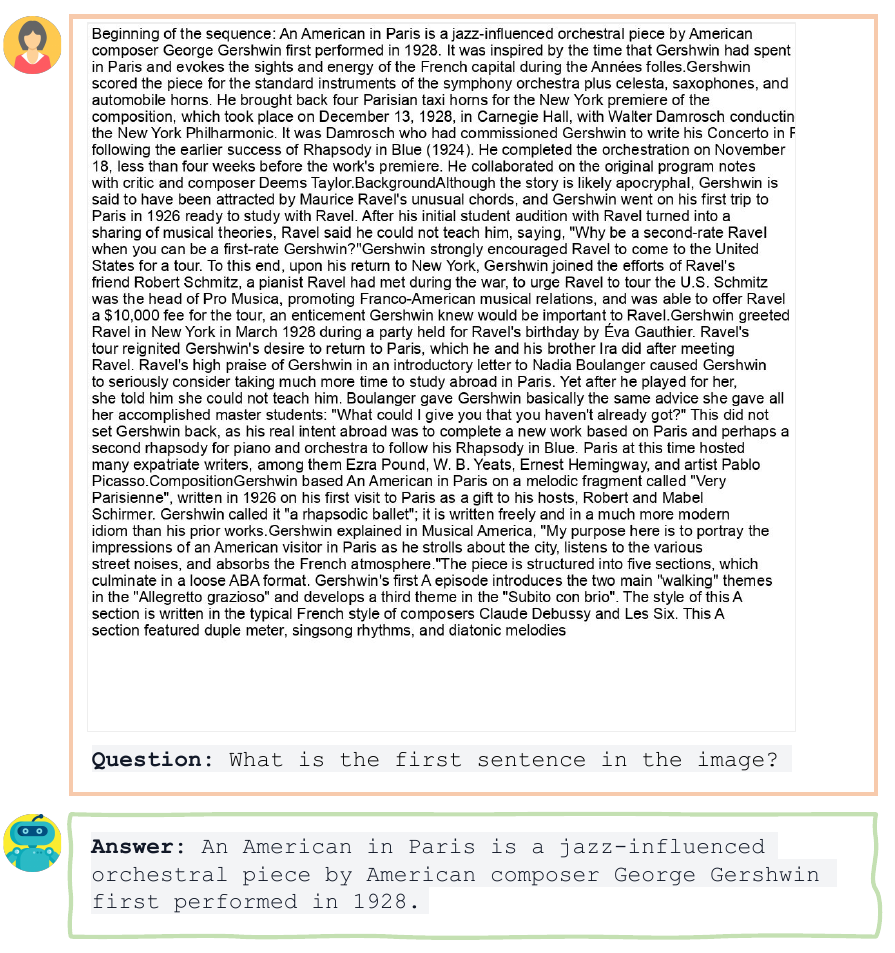}
    \caption{Task \texttt{SentRetrie}.
    }
    \label{fig:task_firstsentence}
\end{figure*}

In Section~\ref{sec:long_context_multimodal_instruct}, we first introduce multimodal long-context tasks categorized in long-form multi-image input and long-form text output. And in Figure~\ref{fig:task_index}-\ref{fig:task_firstsentence}, we visualize full task examples.
\label{sec:longcontext_task_examples}

\clearpage
\section{Discussion}
\subsection{Limitations}

While our model, SEEKER, has made significant strides in processing extended-context multimodal inputs, it encounters several critical limitations that require deeper investigation. The process of compressing textual information into visual tokens, although efficient, may inadvertently overlook precise textual understanding. Future endeavors should focus on developing hybrid encoding strategies that balance token compression with the preservation of essential information. Additionally, SEEKER could inadvertently learn and perpetuate biases present in its training data. It is imperative that further research is conducted to identify, understand, and address these biases, ensuring the model’s equity and inclusiveness.

\subsection{Societal Impact}
By integrating visual tokens with textual data, SEEKER addresses the limitations of traditional models and supports the handling of longer input sequences. This innovation could transform various sectors, improving information accessibility and retrieval systems across academic research, legal document analysis, and extensive data processing tasks. Particularly beneficial in educational and professional environments, SEEKER enables rapid and accurate extraction of vast informational content, fostering better decision-making and knowledge dissemination. However, this advancement might exacerbate information disparities if not equitably accessible. Steps should be taken to make sure it is both affordable and available to everyone.


\end{document}